\documentclass[pdflatex,sn-mathphys-num]{sn-jnl}

\usepackage{graphicx}%
\usepackage{multirow}%
\usepackage{amsmath,amssymb,amsfonts}%
\usepackage{amsthm}%
\usepackage{mathrsfs}%
\usepackage[title]{appendix}%
\usepackage{xcolor}%
\usepackage{textcomp}%
\usepackage{manyfoot}%
\usepackage{booktabs}%
\usepackage{algorithm}%
\usepackage{algorithmicx}%
\usepackage{algpseudocode}%
\usepackage{listings}%
\usepackage{multirow}
\usepackage{subcaption}
\usepackage{pifont} 
\usepackage{dirtree}

\newcommand{\cmark}{\ding{51}}%
\newcommand{\xmark}{\ding{55}}

\theoremstyle{thmstyleone}%
\theoremstyle{thmstyletwo}%

\theoremstyle{thmstylethree}%

\definecolor{codegreen}{rgb}{0,0.6,0}
\definecolor{codegray}{rgb}{0.5,0.5,0.5}
\definecolor{codepurple}{rgb}{0.58,0,0.82}
\definecolor{backcolour}{rgb}{0.95,0.95,0.92}
\lstdefinestyle{mystyle}{
	commentstyle=\color{codegreen},
	keywordstyle=\color{blue},
	numberstyle=\tiny\color{codegray},
	stringstyle=\color{codepurple},
	basicstyle=\ttfamily\small,
	breakatwhitespace=false,         
	breaklines=true,                 
	captionpos=b,                    
	keepspaces=true,                 
	numbersep=5pt,                  
	showspaces=false,                
	showstringspaces=false,
	showtabs=false,                  
	tabsize=2
}

\definecolor{c0}{HTML}{0B6E4F}
\definecolor{c4}{HTML}{870056}
\definecolor{c1}{HTML}{F6830F} 
\definecolor{c2}{HTML}{BB2205} 
\definecolor{c3}{HTML}{1F3C88} 
\definecolor{c5}{HTML}{5B6D02}
\definecolor{c6}{HTML}{600387}

\def\etal{\emph{et al}.}
\begin{document}

\title[Article Title]{PhotoHolmes: a \textit{Python} library for forgery detection in digital images}


\author[1]{\fnm{Julián} \sur{ O'Flaherty}}\email{julian.o.flaherty@fing.edu.uy}
\equalcont{These authors contributed equally to this work.}

\author[1]{\fnm{Rodrigo} \sur{Paganini}}\email{rodrigo.paganini@fing.edu.uy}
\equalcont{These authors contributed equally to this work.}

\author[1]{\fnm{Juan Pablo} \sur{Sotelo}}\email{juan.pablo.sotelo.silva@fing.edu.uy}
\equalcont{These authors contributed equally to this work.}

\author*[1]{\fnm{Julieta} \sur{Umpiérrez}}\email{jumpierrez@fing.edu.uy}
\equalcont{These authors contributed equally to this work.}

\author[2,3]{\fnm{Marina} \sur{Gardella}}\email{marina.gardella@impa.br}

\author[4]{\fnm{Matías} \sur{Tailanián}}\email{matias.tailanian@digitalsense.ai}

\author[1,2]{\fnm{Pablo} \sur{Musé}}\email{pmuse@fing.edu.uy}

\affil*[1]{\orgdiv{IIE, Facultad de Ingeniería}, \orgname{Universidad de la República}, \orgaddress{\street{J. Herrera y Reissig 565}, \city{Montevideo}, \postcode{11300}, \country{Uruguay}}}

\affil[2]{\orgdiv{Centre Borelli}, \orgname{Ecole Normale Supérieure Paris-Saclay}, \orgaddress{\street{4 avenue des Sciences}, \city{Gif-sur-Yvette}, \postcode{91190}, \country{France}}}

\affil[3]{\orgdiv{Centro Pi}, \orgname{Instituto de Matemática Pura e Aplicada}, \orgaddress{\street{Estrada Dona Castorina 110}, \city{Rio de Janeiro}, \postcode{22460-320}, \country{Brazil}}}

\affil[4]{\orgdiv{Digital Sense}, \orgaddress{\street{Brenda 5751}, \city{Montevideo}, \postcode{11400}, \country{Uruguay}}}

\abstract{In this paper, we introduce \textit{PhotoHolmes}, an open-source \textit{Python} library designed to easily run and benchmark forgery detection methods on digital images. The library includes implementations of popular and state-of-the-art methods, dataset integration tools, and evaluation metrics. Utilizing the \textit{Benchmark} tool in \textit{PhotoHolmes}, users can effortlessly compare various methods. This facilitates an accurate and reproducible comparison between their own methods and those in the existing literature. Furthermore, \textit{PhotoHolmes} includes a command-line interface~(CLI) to easily run the methods implemented in the library on any suspicious image. As such, image forgery methods become more accessible to the community.
The library has been built with extensibility and modularity in mind, which makes adding new methods, datasets and metrics to the library a straightforward process. The source code is available at \href{https://github.com/photoholmes/photoholmes}{https://github.com/photoholmes/photoholmes}.}




\maketitle

\section{Introduction}\label{sec1}

Images play a crucial role not only in communications but also in the way we, as humans, perceive the world. There is no room for doubt about the importance of an image when it comes to corroborating a story in the press, as lately, we receive most information through images and videos instead of text. Unfortunately, given the malicious intentions of some individuals and organizations, image forgeries have surfaced, contributing to the negative spiral of \textit{fake news}. At first, most of these forgeries could be detected by the naked eye. However, with the development of powerful image processing software and the convincing results achieved by deep learning techniques, doing the aforementioned analysis is becoming much harder. Therefore, since the seminal work by Popescu and Farid~\cite{popescu_farid}, part of the image processing community has started to develop methods in order to detect forgeries in digital images. 

\begin{figure}[]
    \centering
    \begin{subfigure}{0.29\linewidth}
        \centering
        \includegraphics[width=\linewidth]{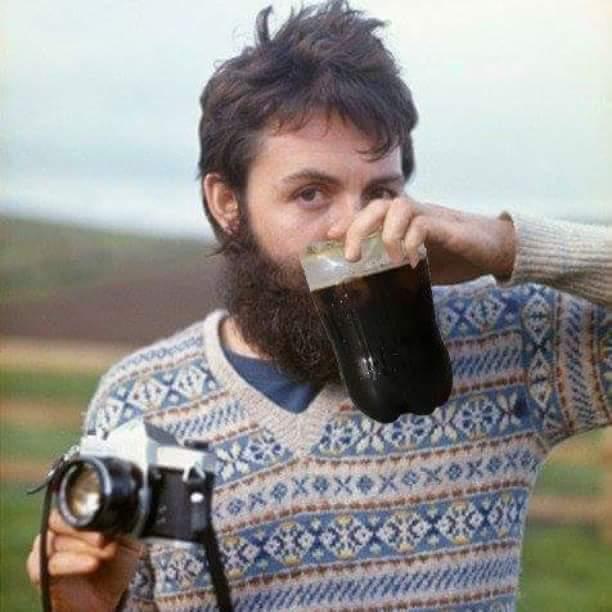}
        \caption{Forged image}
        \label{fig:sub1}
    \end{subfigure}
    \begin{subfigure}{0.29\linewidth}
        \centering
        \includegraphics[width=\linewidth]{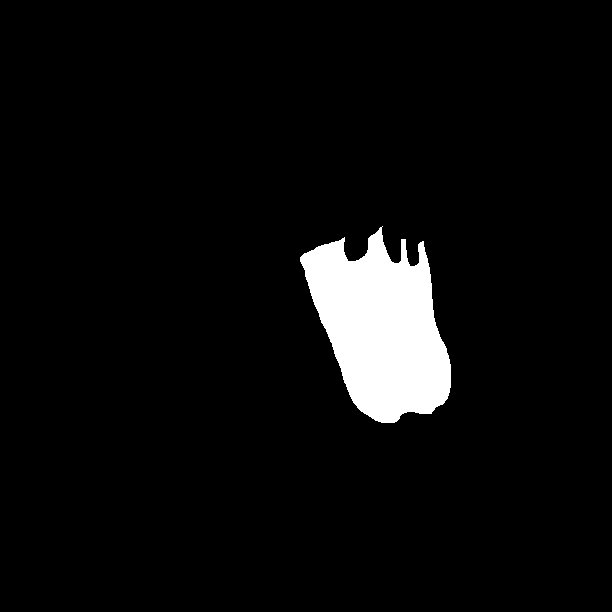}
        \caption{Mask}
        \label{fig:sub2}
    \end{subfigure}
    \begin{subfigure}{0.29\linewidth}
        \centering
        \includegraphics[width=\linewidth]{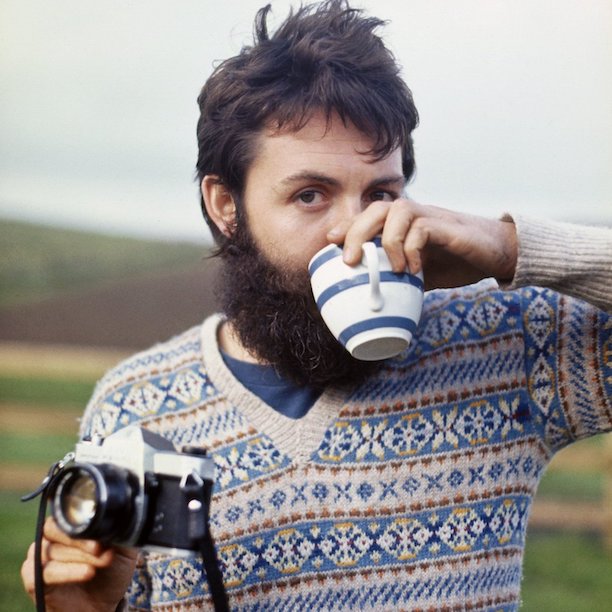}
        \caption{Original image}
        \label{fig:sub3}
    \end{subfigure} \\
    \begin{subfigure}{0.29\linewidth}
        \centering
        \includegraphics[width=\linewidth]{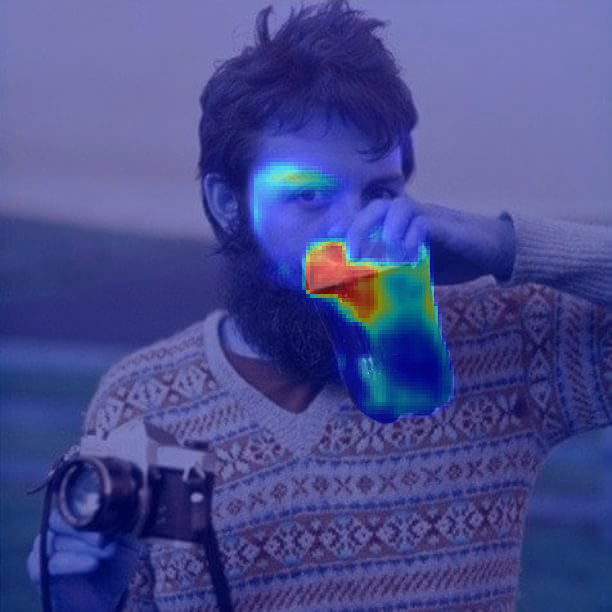}
        \caption{CAT-Net~\cite{catnet}}
        \label{fig:catnet}
    \end{subfigure}
    \begin{subfigure}{0.29\linewidth}
        \centering
        \includegraphics[width=\linewidth]{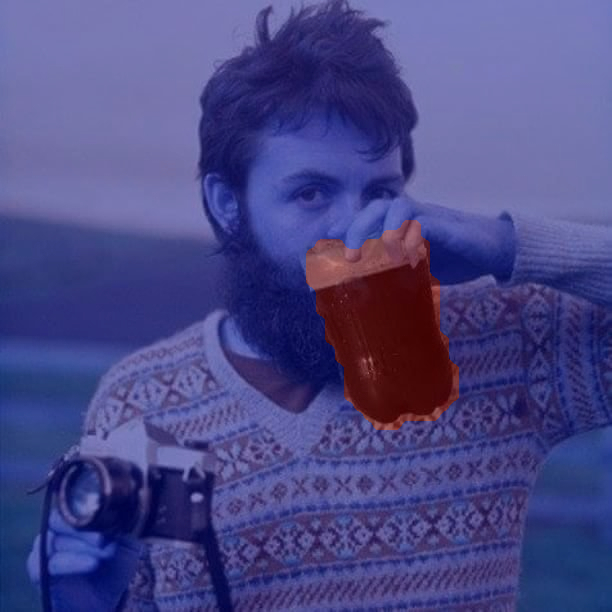}
        \caption{FOCAL~\cite{wu2023rethinking}}
        \label{fig:focal}
    \end{subfigure}
    \begin{subfigure}{0.29\linewidth}
        \centering
        \includegraphics[width=\linewidth]{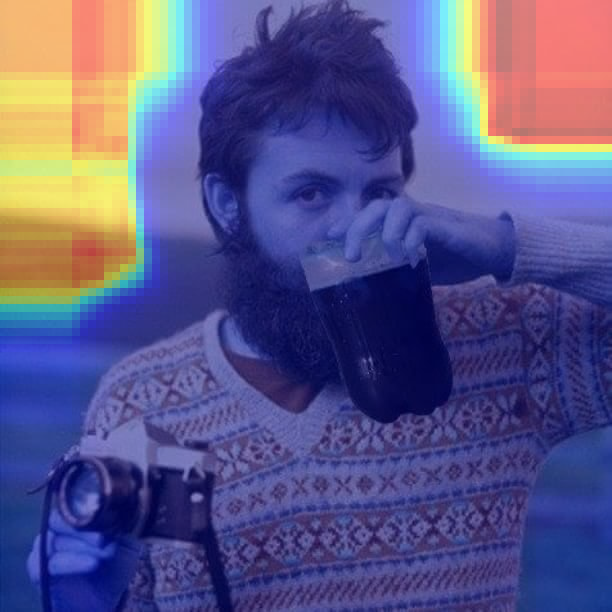}
        \caption{Splicebuster G-U~\cite{splicebuster_cozzolino}}
        \label{fig:splicebustergu}
    \end{subfigure} \\
    \begin{subfigure}{0.29\linewidth}
        \centering
        \includegraphics[width=\linewidth]{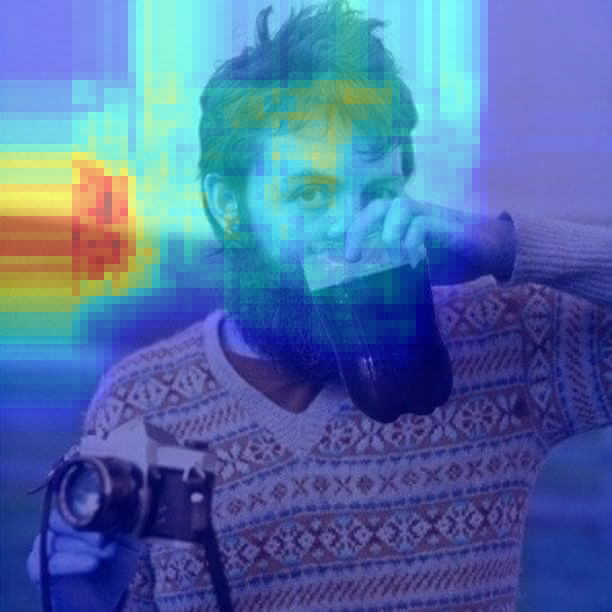}
        \caption{Splicebuster G-G~\cite{splicebuster_cozzolino}}
        \label{fig:splicebustergg}
    \end{subfigure}
    \begin{subfigure}{0.29\linewidth}
        \centering
        \includegraphics[width=\linewidth]{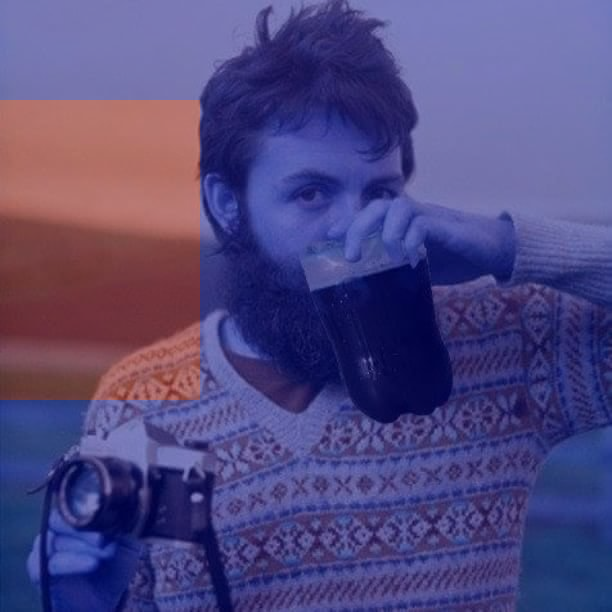}
        \caption{Noisesniffer~\cite{Noisesniffer}}
        \label{fig:sub7}
    \end{subfigure}
    \begin{subfigure}{0.29\linewidth}
        \centering
        \includegraphics[width=\linewidth]{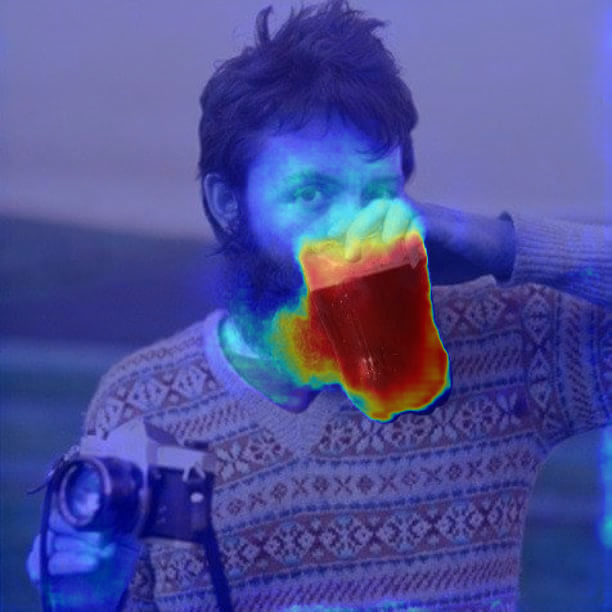}
        \caption{TruFor~\cite{guillaro2023trufor}}
        \label{fig:sub8}
    \end{subfigure}\\
    \begin{subfigure}{0.29\linewidth}
        \centering
        \includegraphics[width=\linewidth]{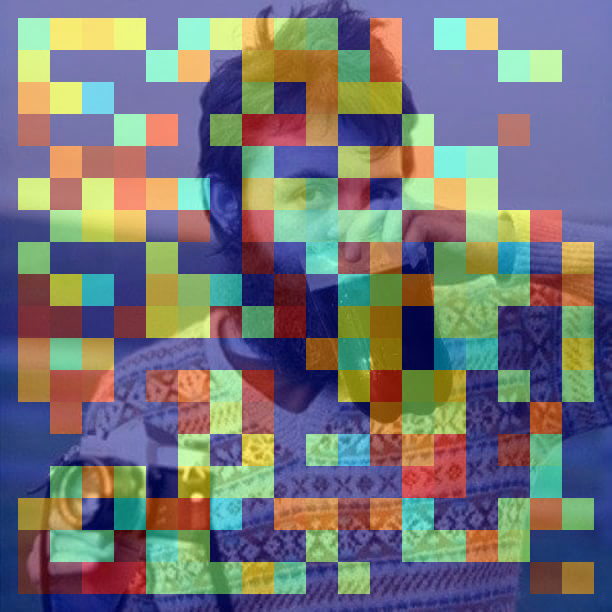}
        \caption{Adaptive CFA~\cite{9157017}}
        \label{fig:adaptive1}
    \end{subfigure} 
    \begin{subfigure}{0.29\linewidth}
        \centering
        \includegraphics[width=\linewidth]{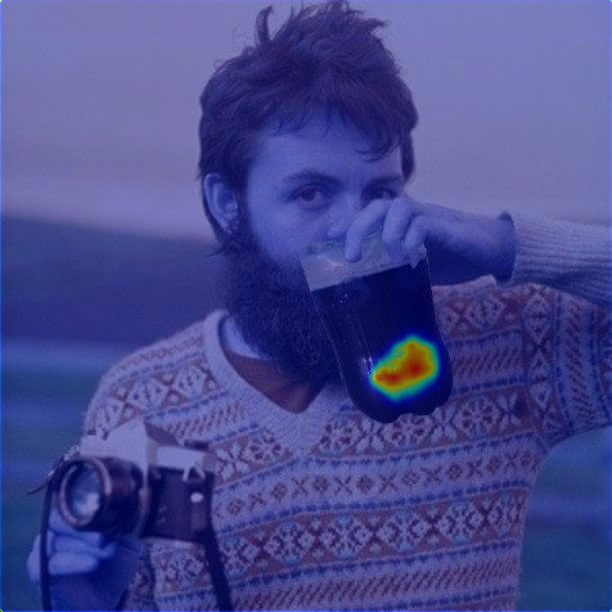}
        \caption{PSCC-Net~\cite{liu2022psccnet}}
        \label{fig:sub11}
    \end{subfigure}
    \begin{subfigure}{0.29\linewidth}
        \centering
        \includegraphics[width=\linewidth]{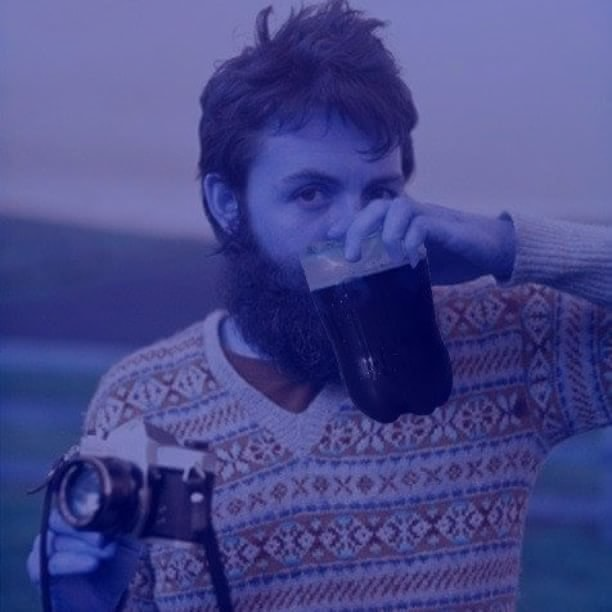}
        \caption{Zero~\cite{zero}}
        \label{fig:sub12}
    \end{subfigure} \\
    \begin{subfigure}{0.29\linewidth}
        \centering
        \includegraphics[width=\linewidth]{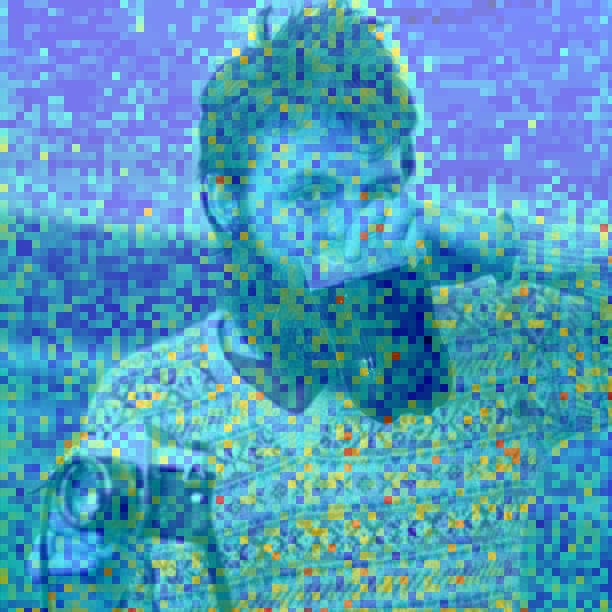}
        \caption{DQ~\cite{dq_lin}}
        \label{fig:sub13}
    \end{subfigure}
    \begin{subfigure}{0.29\linewidth}
        \centering
        \includegraphics[width=\linewidth]{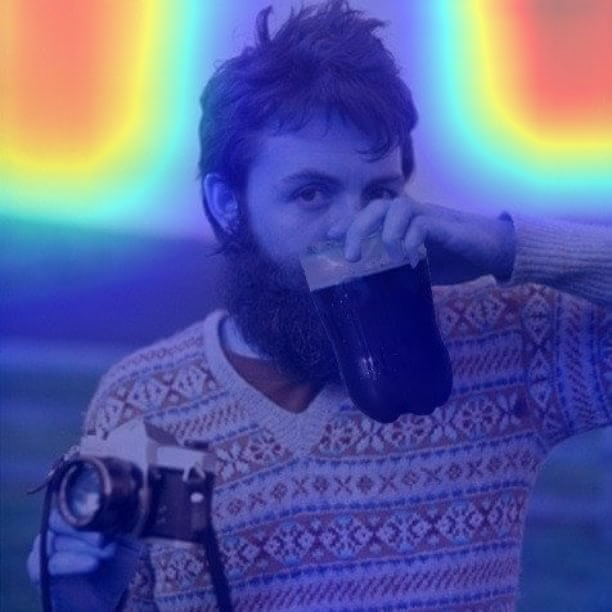}
        \caption{EXIF MS~\cite{zheng2023exif}}
        \label{fig:sub14}
    \end{subfigure}
    \begin{subfigure}{0.29\linewidth}
        \centering
        \includegraphics[width=\linewidth]{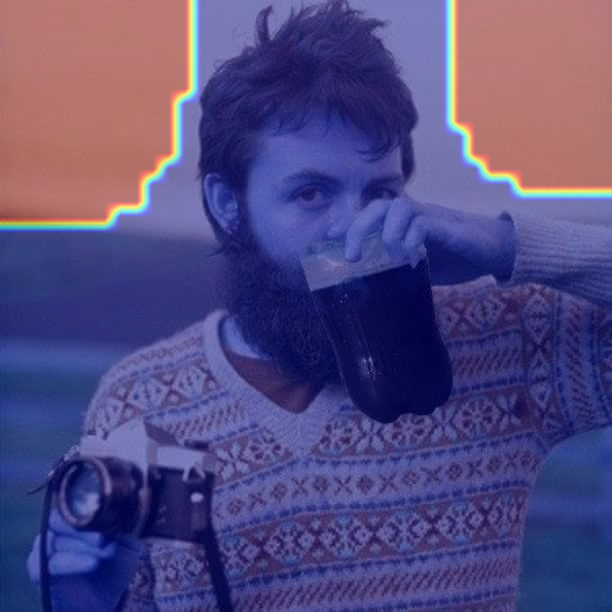}
        \caption{EXIF Ncuts~\cite{zheng2023exif}}
        \label{fig:sub15}
    \end{subfigure}
    \vspace*{-0.2cm}
    \caption{Results of all the methods implemented in the first version of \textit{PhotoHolmes} on a satirical image of Paul McCartney drinking fernet spread on social networks (Figure~\ref{fig:sub1}). For Splicebuster~\cite{splicebuster_cozzolino}, we include, the result with Gaussian-Uniform (Figure~\ref{fig:splicebustergu}) and Gaussian-Gaussian for the EM (Figure~\ref{fig:splicebustergg}). As for EXIF~\cite{zheng2023exif}, we included both the result using mean shift (Figure~\ref{fig:sub14}) and the one with normalized cuts as clustering method (Figure~\ref{fig:sub15}).} 
    \label{fig:outputs}
    \vspace*{-0.3cm}
\end{figure}

The increasing number of publications in image forgery detection highlights the need for a standardized library integrating methods, metrics, and datasets. Ideally, such a library should allow both to benchmark methods in popular forgery datasets and test suspicious images using different methods without having to go through each of the methods' implementations separately. Considering most methods have strengths and weaknesses related to the forgery type, being able to run an image through different methods with ease enables the quick creation of robust forgery detection reports on a suspicious image. An example is illustrated in Figure~\ref{fig:outputs} that shows a satiric photomontage of Paul McCartney spread on social networks with all the results given by \textit{PhotoHolmes}. 

With these considerations in mind, we created \textit{PhotoHolmes}. This novel \textit{Python} library provides different modules allowing users to run and benchmark state-of-the-art methods. Furthermore, the library is designed to be modular, reproducible, and extensible, ensuring the ability to easily add new methods as they are published and compare them with the existing ones. 
The objective is to contribute to the forgery detection community by making it easier to benchmark methods and test suspicious images. 

The goal of this paper is to present and describe the \textit{PhotoHolmes} library. The rest of the paper is organized as follows: Section~\ref{sec:related_works} summarizes related works. Section~\ref{sec:design_principles} describes the considered design principles. 
Then, Section~\ref{sec:photoholmes} provides a detailed description of the library modules. Section~\ref{sec:use_cases} presents examples of the two main features of \textit{PhotoHolmes}: the benchmarking module and the ability to run all methods on a suspicious image. Conclusions and a summary of future work are presented in Section~\ref{sec:conclusions}. 

\section{Related works}
\label{sec:related_works}

Related to \textit{PhotoHolmes}, several projects have emerged during the past few years with the goal of unifying the state-of-the-art in image forgery detection, making it easier to use by both, academics and the general public. The most related to ours is the \textit{Matlab} toolkit~\cite{zampoglou2017large} introduced in 2017. This toolbox includes the implementation of several algorithms that were considered state-of-the-art at that time. Still, a limitation of their work is its reliance on \textit{Matlab}, which is proprietary software. This does not favor usability and also makes it difficult to build a community that contributes with new methods. Adding to this, the fact that all of the state-of-the-art methods are implemented in \textit{Python} explains why it has not been updated lately. Related to that work is the Image Verification Assistant~\cite{itiMeVerImage}, a website created by the same authors that allows users to upload an image and returns the results obtained with several algorithms. Though this website allows the general public to test their own suspicious images, it is not well-suited for benchmark purposes. Besides, the code of the implemented methods is not publicly available.

The last remarkable related work is the \emph{InVID plugin}~\cite{invidprojectInVIDVerification}, a browser plugin developed by the \textit{Agence France Presse} (AFP) with the aim of helping journalists verify information coming from social networks. Amongst several functionalities, the plugin provides forensic analysis of suspicious images by displaying the results obtained by different algorithms on the image to be tested. This plugin, which is public and free, can be used directly from a browser. This allows people without coding expertise to be able to test images easily. Still, as in the case of the Image Verification Assistant~\cite{itiMeVerImage}, such platforms are not well-suited for benchmark purposes and, again, the source code of the methods is not publicly available.

Despite the timid attempts in the field of forgery detection to develop a unified open-source library, such kinds of libraries have already emerged in other fields. A notable example is the \textit{anomalib} library~\cite{akcay2022anomalib}, a \textit{Python} library specially designed to benchmark and develop anomaly detection methods. Even though the said library was designed with another problem in mind, the core design principles lay really close to ours. Other remarkable examples are the libraries developed by \mbox{\textit{OpenMMLab}}. These include \textit{MMSegmentation}~\cite{mmseg2020} for image segmentation, \textit{MMPose}~\cite{mmpose2020} for pose estimation, \textit{MMOCR}~\cite{mmocr2021} for text detection, recognition and understanding and \textit{MMDetection}~\cite{mmdetection} for object detection, just to mention a few. 
Our contribution is to provide a centralized, open-source and extensible library to allow researchers to develop and compare their methods in 
a reproducible and standardized manner.

\section{Design principles and design choices}
\label{sec:design_principles}

To ensure the library fulfills its purpose and is maintained over time, the library is designed with four main principles: modularity, reproducibility, extensibility, and usability.

\vspace{0.2cm}
\noindent\textbf{Modularity.} The library is designed to be modular, where each module tackles a specific aspect of the image forgery detection pipeline. This simplifies maintenance work and makes for cleaner and simpler code.

\vspace{0.2cm}
\noindent\textbf{Reproducibility.} The library is designed to ensure reproducibility, enabling the user to easily replicate the experimental results of the different methods. Furthermore, it provides a transparent open-source implementation and detailed documentation of the methods. 

\vspace{0.2cm}
\noindent\textbf{Extensibility.} The library is designed to be extensible, allowing contributors to quickly expand the functionalities of the library by adding new methods, datasets, metrics, and tools. This flexibility is facilitated through a wrapper-based code architecture complemented by detailed guides on integrating new functionalities.

\vspace{0.2cm}
\noindent\textbf{Usability.} The library is designed to be easy to use. This is accomplished through a Command Line Interface (CLI), which allows the user to invoke the library without writing code. 
Additionally, our modules and classes are designed for seamless integration of the library into the user's code.
\\

Following the aforementioned principles, we designed and wrote the library following an Object Oriented Programming (OOP) paradigm. In OOP, one can define a parent class that sets a basic structure to follow, from which new classes can be defined, inheriting properties and methods. The second reason OOP was chosen is because the language chosen, \emph{Python}~\cite{python}, is an interpreted OOP programming language.

Choosing \emph{Python} as the programming language has its roots in the popularity it has gained in the computer vision and data science community in general, coming to a point where most of the research carried out today is written in 
this language.
\emph{Python}'s 
strong foothold in the data science community comes from its simple syntax and its extensive support of third-party libraries. Libraries such as \emph{NumPy}~\cite{harris2020array} and \emph{PyTorch}~\cite{Ansel_PyTorch_2_Faster_2024} enable quick and efficient numerical computations, allowing it to reach speeds close to a compiled language like \textit{C}, but with a much simpler syntax.

The last important design choice refers to the deep learning framework used. There are many deep learning frameworks, but there are two that stand out: \emph{Tensorflow}~\cite{Abadi_TensorFlow_Large-scale_machine_2015} and \emph{PyTorch}~\cite{Ansel_PyTorch_2_Faster_2024}. Until a few years ago, \emph{Tensorflow} was the most widely used deep learning framework, and some image forensics methods were implemented using it. Lately, the tide has turned in favor of \emph{PyTorch}~\cite{paperswithcode-trends}, and most recent research has been trained using this framework, which is why we chose only to support \emph{PyTorch} in this first version.

\section{The PhotoHolmes library}
\label{sec:photoholmes}

The library is subdivided into 7 different modules, each of them with a specific purpose. The modules are the following:
\begin{itemize}
    \item \textbf{Datasets}: contains the code implementation for loading the different datasets available for benchmarking the methods.
    \item \textbf{Preprocessing}: contains different preprocessing operations that can be applied to the images before using the methods.
    \item \textbf{Methods}: contains the methods that can be used to detect forgeries. 
    \item \textbf{Postprocessing}: contains different postprocessing functions that can be used to post-process the outputs of the methods.
    \item \textbf{Metrics}: contains the different metrics useful for evaluating the performance of the methods.
    \item \textbf{Benchmark}: contains the benchmark class that allows the user to benchmark a method with a list of metrics in different datasets.
    \item \textbf{CLI}: contains the CLI that allows the user to use the library from the command line.
\end{itemize}

Following the modularity principle, each module tackles a specific aspect of the forgery detection pipeline. The \emph{Datasets}, \emph{Preprocessing}, \emph{Methods}, and \emph{Metrics} are designed to work in unison, but each of the modules can be used independently from each other. The \emph{Postprocessing} module groups useful functions used at the end of a method's pipeline, so the \emph{Methods} module depends on it. The \emph{Benchmark} and \emph{CLI} modules are both designed to run image forgery pipelines, so naturally, they both make use of all previously mentioned modules.

\subsection{Datasets} \label{subsec:datasets}

The \emph{Datasets} module contains a compilation of popular datasets that are used to evaluate methods. In the library, a Dataset is a class with instructions to locate and load the images of a dataset.

Following OOP principles, we define a \emph{BaseDataset} to which we add common attributes and methods that any dataset might use. In particular, the data loading logic is implemented in a way that is reusable for all datasets, needing only to override simple properties and methods to define the folder structure. Some methods, like the ones that define the folder structure, need to be overwritten when creating a new dataset, while others should be overwritten when the default implementation is not fit for the dataset, like mask binarization. An important note is that the \mbox{\emph{BaseDataset}} inherits from \emph{PyTorch}'s dataset, meaning any \emph{PhotoHolmes}' dataset can be used within that framework.

We identified three types of image data that methods use: the pixel values of the image itself, its DCT coefficients, and the quantization tables (\texttt{qtables}) of JPEG images. With that in mind, our datasets can load the three types of data, specified either through the \texttt{load} parameter or by the preprocessing pipeline's input, which will be defined in the next section. As more image forgery methods are developed, support for more image data can and will be added.

Another important parameter is \texttt{tampered\_only}, which allows the user to specify whether they want all the images to be loaded or only those where a forgery does exist.\footnote{This setting is useful when evaluating forgery localization.}.\


The first release of \emph{PhotoHolmes} includes 7 benchmarking datasets: Columbia~\mbox{\cite{columbia_dataset, Wu2022}}, CASIA 1.0~\cite{CASIA-V1, CASIA-v1-gt, Wu2022}, DSO-1~\cite{DSO1, Wu2022}, Korus~\cite{Korus2016TIFS, Korus2016WIFS}, AutoSplice~\cite{jia2023autosplice} and Trace~\cite{bammey2021nonsemantic}. On top of the original versions, we include the social media versions of Columbia, DSO-1, and CASIA v1, as well as WebP compressed versions of Korus and Columbia. The selected datasets cover a wide range of forgery types and image formats, which we deemed important to benchmark the diverse array of included methods. The most relevant information about the included datasets is summarized in Table~\ref{tab:popular_datasets}. A short description of each of them can be found in Appendix~\ref{app:datasets}.

\begin{table*}[]
\resizebox{\textwidth}{!}{%
\begin{tabular}{lccccc}
\toprule
Dataset & Types of forgery   & Nb. of images (\textcolor{blue}{forged} + \textcolor{codegreen}{pristine}) & Format  & SM version & WebP version\\ 
\midrule
\multirow{2}{*}{Columbia~\cite{columbia_dataset, Wu2022}}      & \multirow{2}{*}{Splicing}                    & \multirow{2}{*}{363 (\textcolor{blue}{180} + \textcolor{codegreen}{183})}                  & \multirow{2}{*}{TIF} & \multirow{2}{*}{\cmark}                   &       \multirow{2}{*}{\cmark}   \\
& & & & &\\
Coverage~\cite{wen2016}         & Copy-move                   & 200    (\textcolor{blue}{100} + \textcolor{codegreen}{100}) & TIF & \xmark                   & \xmark                                      \\ [10pt] 
DSO-1~\cite{DSO1, Wu2022}         & Splicing                    & 200 (\textcolor{blue}{100} + \textcolor{codegreen}{100})  & PNG                  & \cmark                 & \xmark                                       \\ [10pt] 

\multirow{2}{*}{Korus~\cite{Korus2016TIFS, Korus2016WIFS}} & Splicing, copy-move & \multirow{2}{*}{440 (\textcolor{blue}{220} + \textcolor{codegreen}{220})}  & \multirow{2}{*}{TIF}                 & \multirow{2}{*}{\xmark}                   & \multirow{2}{*}{\cmark}                                      \\ 
 & object removal &  &  &  &   \\ [10pt] 
Casia 1.0~\cite{CASIA-V1, CASIA-v1-gt, Wu2022}      & Splicing, copy-move      & 1023 (\textcolor{blue}{923} + \textcolor{codegreen}{100})             & JPEG    &    \cmark      &   \xmark                                    \\[10pt] 
AutoSplice~\cite{jia2023autosplice}       & Generative inpainting       & 5894  (\textcolor{blue}{3621} + \textcolor{codegreen}{2273})                & JPEG  & \xmark  & \xmark                  \\ [10pt] 
\multirow{2}{*}{Trace~\cite{bammey2021nonsemantic}}       & Alterations to        & \multirow{2}{*}{24000 (\textcolor{blue}{24000} + \textcolor{codegreen}{0})}         & \multirow{2}{*}{PNG}          & \multirow{2}{*}{\xmark} & \multirow{2}{*}{\xmark}               \\ 
& acquisition pipeline       &  &  & &  \\ 
\bottomrule
\end{tabular}%
}
\caption{Summary of the main characteristics of the datasets included in the first release of \textit{PhotoHolmes}, such as the type of forgery they feature, the number of images (both pristine and forged) included in each of them, the images' format and whether their social media (SM) and WebP versions are also incorporated.}
\label{tab:popular_datasets}
\end{table*}

Following the design principles on which we built \emph{PhotoHolmes}, using the included datasets is a straightforward process. For example, to get and plot the first image of the Columbia dataset, the following code snippet can be used:

\lstset{style=mystyle}

\begin{lstlisting}[language=Python]
from photoholmes.datasets.columbia import ColumbiaDataset
from photoholmes.utils.image import plot

# Load the dataset
dataset_path = "data/Columbia"
dataset = ColumbiaDataset(
    dataset_path=dataset_path,
    preprocessing_pipeline=None,
    tampered_only=True,
    load=["image"]
)

# Get the first image
data, mask, image_name = dataset[0]
image = data["image"]
plot(image)
\end{lstlisting}

In addition to the dataset definitions, the module contains a registry that lists the available datasets and a factory that allows the user to easily load any of them.
Given the simple extensibility achieved with how the module was designed, we expect to continue growing the \emph{PhotoHolmes} dataset registry as new datasets are proposed.

\subsection{Preprocessing}

Most forgery detection methods work on transformations of the image data rather than on the image itself. Some transformations are simple, for example, a grayscale transformation, while others require more complex operations like computing the DCT volumes in CAT-Net~\cite{catnet}. To give structure to these transformations, we define the \emph{Preprocessing} module. 

Within this module, we define a \texttt{BasePreprocessing} class. This class handles the application of preprocessing operations to the data, only requiring children classes to implement the \texttt{\_\_call\_\_} method. In order to allow preprocessing operations to be mixed and matched in different pipelines, each preprocessing operation expects a dictionary as input and outputs a dictionary. The preprocessing operations can modify, add, or remove entries in this dictionary as long as they are composed in a compatible fashion. In the first release of \emph{PhotoHolmes}, we include the following preprocessing operations:
\begin{itemize}
    \item \texttt{ZeroOneRange}: changes the image pixel values from $[0,255]$ to $[0,1]$.
    \item \texttt{Normalize}: applies standardization to the image by subtracting the mean and dividing by the standard deviation.
    \item \texttt{RGBtoGray}: converts an image from the RGB colorspace to grayscale.
    \item \texttt{GraytoRGB}: converts the image from grayscale to RGB.
    \item \texttt{RoundToUInt}: rounds the input float tensor and converts it to an unsigned integer.
    \item \texttt{ToNumpy}: converts tensors to \emph{numpy} arrays.
    \item \texttt{ToTensor}: converts a \emph{numpy} array to a \emph{Torch} tensor.
    \item \texttt{GetImageSize}: adds the size of the image to the dictionary. 
\end{itemize}
Some methods require preprocessing operations outside this list, but given their specificity, we opted to define them within the method's module. Each method has a \texttt{preprocessing.py} file where custom preprocessing operations can be defined, and more importantly, the preprocessing pipeline is defined.

The \texttt{PreprocessingPipeline} is a class that sequentially runs a list of preprocessing operations on the input data and leaves it ready for the method to intake. The pipeline is designed to be easy to use, simplifying the composition of transforms and controlling the input to the model. The \texttt{PreprocessingPipeline} can be given to a \emph{Dataset} as a parameter to apply the transformations when loading the data. The \texttt{PreprocessingPipeline} requires three parameters: the list of preprocessing operations, the input keys that need to be included in the initial dictionary provided to the pipeline, and the output keys that the method expects. The input keys are used to validate the pipeline input, avoiding errors when used incorrectly, and are also used by the \emph{Datasets} module to load only the necessary image information. The output keys are used to filter out any extra keys that were left over during the preprocessing operations.

Here is a code snippet that implements a simple preprocessing pipeline. It expects an image, converts it to a \emph{numpy} array, and then converts it to grayscale.

\lstset{style=mystyle}
\begin{lstlisting}[language=Python]
from photoholmes.preprocessing import ToNumpy, RGBtoGray, PreProcessingPipeline
from photoholmes.utils.image import read_image

pipeline = PreProcessingPipeline(
    transforms=[ToNumpy(image_keys=["image"]), RGBtoGray()],
    inputs=["image"],
    outputs_keys=["image"]
)

image = read_image("example_image.jpeg")
result = pipeline(image=image)
\end{lstlisting}

\subsection{Methods} \label{subsec:methods}

The \emph{Methods} module is the core of the library, and all the modules are designed around it. The implementations of forgery detection methods are diverse in code structure, programming style, inputs, outputs, and documentation, making it difficult to run quick inference or evaluation unless the authors provide specific scripts for it. To address this issue, following OOP principles, we designed a \emph{BaseMethod} class that all methods inherit from, and that ensures compatibility with the rest of the modules provided in the library. Additionally, we defined \emph{BaseTorchMethod}, meant to be used by those methods that rely solely on neural networks. This way, the methods are compatible with \emph{PyTorch} and can be used for retraining and other experiments. 

As with the \emph{Datasets} module, the \emph{BaseMethod} includes default implementations of some functionalities like loading from a configuration file, and it requires the user to implement the \texttt{benchmark} function.\footnote{This function is, in fact, a method of the object. For clarity, we chose not to use this term since we were discussing forgery detection methods.}
This function will be used by the \emph{Benchmark} module we will later introduce, simplifying a method's evaluation process. Another notable function\footnote{This is also a method of the object. As before, for clarity, we avoid this term here.} is \texttt{to\_device} 
that allows the user to move the method's computation into another device, such as GPU, which is commonly used to accelerate inference time in deep learning models.

This first version of the \textit{PhotoHolmes} library includes ten state-of-the-art methods: Adaptative CFA~\cite{9157017}, Noisesniffer~\cite{Noisesniffer}, Zero~\cite{zero}, DQ~\cite{dq_lin}, CAT-Net~\cite{catnet}, Splicebuster~\cite{splicebuster_cozzolino}, EXIF as  Language~\cite{zheng2023exif}, PSCC-Net~\cite{liu2022psccnet}, TruFor~\cite{guillaro2023trufor} and FOCAL~\cite{wu2023rethinking}. Such methods were not only chosen for their performance but also because of their complementarity. Indeed, while some of these methods search for inconsistencies in specific traces of the image processing~\cite{9157017, Noisesniffer, zero, catnet}, others can be regarded as multi-purpose tools that detect inconsistencies from multiple traces simultaneously~\cite{splicebuster_cozzolino, zheng2023exif, liu2022psccnet, guillaro2023trufor, wu2023rethinking}. Table~\ref{tab:methods_summary} summarizes the techniques used by each method and the kind of output they provide. A short description of each of them can be found in Appendix~\ref{app:methods}. 

\begin{table}[]
\resizebox{0.8\columnwidth}{!}{%
\begin{tabular}{lccccc}
\toprule
\multirow{ 2}{*}{Method}  & Target & Deep-
& \multicolumn{3}{c}{Outputs} \\
 & traces & learning & Heatmap & Mask & Detection \\
\midrule
 
\multirow{2}{*}{Adaptive CFA~\cite{9157017}}     & \multirow{2}{*}{CFA} & \multirow{2}{*}{\cmark}  & \multirow{2}{*}{\cmark}           & \multirow{2}{*}{\xmark}        & \multirow{2}{*}{\xmark}             \\
 &  &  &  & &  \\

Noisesniffer~\cite{Noisesniffer}     & Noise   & \xmark      & \xmark           & \cmark        & \cmark             \\[10pt]  
Zero~\cite{zero}             & JPEG    & \xmark      & \xmark           & \cmark        & \cmark             \\[10pt]  
DQ~\cite{dq_lin}               & JPEG    & \xmark      & \cmark           & \xmark        & \xmark             \\[10pt]  
CAT-Net~\cite{catnet}           & JPEG    & \cmark      & \cmark           & \xmark        & \xmark             \\ [10pt] 
Splicebuster~\cite{splicebuster_cozzolino}     & Multiple & \xmark & \cmark           & \xmark        & \xmark             \\[10pt] 
EXIF~\cite{zheng2023exif} & Multiple & \cmark & \cmark           & \cmark        & \cmark             \\[10pt]  
PSCC-Net~\cite{liu2022psccnet}          & Multiple & \cmark & \cmark           & \xmark        & \cmark             \\[10pt]  
TruFor~\cite{guillaro2023trufor}           & Multiple & \cmark & \cmark           & \xmark        & \cmark             \\[10pt]  
FOCAL~\cite{wu2023rethinking}            & Multiple & \cmark & \xmark           & \cmark        & \xmark             \\ 
\bottomrule
\end{tabular}%
}
\caption{Summary of the techniques used by each of the methods included in the first release of \emph{PhotoHolmes} as well as the kind of output they provide. The outputs can be continuous heatmaps representing a probability, binary masks and detection scores.}
\label{tab:methods_summary}
\end{table}

As an example of how the \textit{Methods} module can be used, we provide a code snippet to run CAT-Net on an image. In a few lines of code, we instantiate the method, change the device to GPU for a faster inference, and run the method on the image.

\lstset{style=mystyle}

\begin{lstlisting}[language=Python]
from photoholmes.methods.catnet import CatNet, catnet_preprocessing
from photoholmes.utils.image import read_image, read_jpeg_data

path_to_image = "path_to_image"
image = read_image(path_to_image)
dct, qtables = read_jpeg_data(path_to_image)

# Preprocess data
image_data = {"image": image, "dct_coefficients": dct, "qtables": qtables}
input = catnet_preprocessing(**image_data)

# Declare the method and use .to_device if you want to run it on cuda or mps instead of cpu
arch_config = "pretrained"
path_to_weights = "path_to_weights"
method = CatNet(
    arch_config=arch_config,
    weights=path_to_weights,
)
device = "cuda"
method.to_device(device)

# Use predict to get the final result
output = method.predict(**input)
\end{lstlisting}

Just as in the \emph{Datasets} module, the \emph{Methods} module includes both a factory and a registry that simplifies loading a model. The registry contains a list of all the available models, while the factory loads the method and associated preprocessing pipeline. Newer releases of \emph{PhotoHolmes} will include more methods according to the demand of the forgery detection community.

Not all the methods included in \emph{PhotoHolmes} have a commercial license. In an effort to include as many methods as possible while respecting the original author's rights, the decision was reached to include the original license inside the method's folder and to log warning messages advising the user to check whether the specific method's license is within their scope of use.

\subsection{Postprocessing}

Postprocessing is a common step in image forgery detection, as many methods employ sliding window predictions or other sub-sampling strategies that yield a prediction smaller than the input image or have an output that has to be rescaled to a different dynamic range. Given that most methods employ at least one of these functions, a module was created to centralize them for reusability.

Another commonly applied postprocessing is casting types and moving data across devices (i.e., GPU to CPU). While the method itself might not need this for prediction, they are useful to integrate with other parts of the \textit{PhotoHolmes} library or even third-party libraries.

Having identified the two main uses of postprocessing, the first version of \textit{PhotoHolmes} includes the following postprocessing functions:
\begin{itemize}
    \item \texttt{to\_device\_dict}: moves dictionary values to the specified device.
    \item \texttt{to\_tensor\_dict}: converts dictionary values to tensors.
    \item \texttt{to\_numpy\_dict}: converts dictionary values to \emph{numpy} arrays.
    \item \texttt{zero\_one\_range}: rescales the output to $[0,1]$.
    \item \texttt{resize\_heatmap\_with\_trim\_and\_pad}: zero-pads or trims the heatmap to match the original image size.
    \item \texttt{upscale\_mask} and \texttt{simple\_upscale\_heatmap}: interpolates the mask or heatmap to match the original image size. 
\end{itemize}

Unlike the previously introduced modules, \emph{Postprocessing} does not have a base class that gives structure. This decision was taken to simplify the image forgery pipeline, choosing to have the method output a comparison-ready mask rather than having to include instructions or a pipeline to transform the output. 
However, some methods may need custom postprocessing, as is the case for Splicebuster. In these cases, there are no restrictions regarding the structure of the postprocessing, but we choose to have the code inside a \texttt{postprocessing.py} file.

\subsection{Metrics} \label{subsec:metrics}

The \textit{Metrics} module is crucial in a forgery detection library, like \textit{PhotoHolmes}, because it provides a standardized way to evaluate the performance of different detection methods. This standardization is essential for comparing results accurately and ensuring reproducible results in research. By implementing this module, \textit{PhotoHolmes} allows users to consistently assess the effectiveness of various algorithms, leading to clearer insights and more reliable conclusions.

The module contains different metrics that are divided into two categories: Metrics imported from \textit{Torchmetrics}~\cite{torchmetrics} and custom metrics that inherit from the \textit{Torchmetrics} base class and implement the metric. The module also contains a registry that lists the different metrics and a factory that allows the user to easily select the implemented metrics they want to use.

The metrics imported from \textit{Torchmetrics} that are included in the first release of \textit{PhotoHolmes} are: 
the Receiver Operating Characteristic curve (ROC) and the area under it (AUROC), the True Positive Rate (TPR), the Intersection over Union (IoU), the  Matthews Correlation Coefficient (MCC), the Precision and the F1 score. If the predicted tensor is a float tensor, which means the prediction is a heatmap, we use the predefined \textit{Torchmetrics} threshold of $0.5$.
These metrics are included by the use of wrappers, and if more \textit{Torchmetrics} metrics are useful for evaluation, they could easily be imported into the \textit{PhotoHolmes} library.

In addition, we also included custom metrics in the first release of \textit{PhotoHolmes}. These metrics are the False Positive Rate (FPR), the weighted IoU$_w$, F1$_w$, and MCC$_w$, 
in two versions: $v_1$ and $v_2$. 

Weighted metrics were introduced by Gardella~\etal~\cite{Noisesniffer} and Bammey~\etal~\cite{bammey2021nonsemantic} as a way of comparing methods whose outputs are heatmaps with methods delivering binary masks. To compute such metrics, we weight the confusion matrix using the heatmap, which corresponds to considering that the pixel value is its probability of being forged. When it comes to the detection task, the weighted scores 
are achieved by weighting
the TP, TN, FP, and FN according to the detection score given by the method.

In addition, when evaluating a method over a full dataset, there are different ways of computing such metrics. The first version ($v_1$) corresponds to calculating the corresponding metric for each pair of prediction and target and then averaging those values to obtain the final score. To compute the second version ($v_2$), we accumulate the false positives, false negatives, true positives, and true negatives over the full dataset and then compute the final weighted metric. 
To understand better the purpose of each metric, 
we split the problem into two tasks: \emph{detection} and \emph{localization}.
This means, whether we aim to detect if a forgery is present or whether we wish to localize where it is in the image.
It is suggested to use the first version ($v_1$) to evaluate the localization problem, as it better captures how the methods perform in each image, while the second version ($v_2$) should be used for the detection problem since the ($v_1$) does not make sense when the output is a single number.

The newly defined weighted metrics behave the same as the standard metrics, as the only change comes in the definition of TP, TN, FP and FN. In our review of the state of the art~\cite{tesis_photoholmes} using \textit{PhotoHolmes}, we selected MCC, F1, and IoU to evaluate localization. MCC is particularly effective in penalizing overestimation of forged areas, stemming from the use of both FP and FN in its calculation. 

A summary of the metrics included in the first release of \textit{PhotoHolmes} is provided in Table~\ref{tab:metrics}, where we classified the metrics according to their weighted or non-weighted nature and the way in which they are computed over the full dataset. For a detailed description of said metrics, we refer the reader to the Appendix~\ref{app:metrics}.

\begin{table}[]
\resizebox{\columnwidth}{!}{%
\bgroup
\begin{tabular}{lcc}
\toprule
& Non-weighted & Weighted \\
\midrule
\multirow{2}{*}{Dataset-level score} & ROC, AUROC, TPR, FPR & \multirow{2}{*}{mAUROC} \\
 & Precision, F1, IoU, MCC & \\[3pt]
Average image-level score & F1$^{v_2}_w$, IoU$^{v_2}_w$, MCC$^{v_2}_w$ & F1$^{v_1}_w$, IoU$^{v_1}_w$ and MCC$^{v_1}_w$\\
\bottomrule
\end{tabular}
\egroup
}
\caption{Summary of metrics included in the first release of \textit{PhotoHolmes}. The ROC, AUROC, TPR, IoU, MCC, Precision, and F1 metrics are imported from \textit{Torchmetrics}. The FPR, F1$^{v_2}_w$, IoU$^{v_2}_w$, MCC$^{v_2}_w$, F1$^{v_1}_w$, IoU$^{v_1}_w$ and MCC$^{v_1}_w$ metrics are custom.}
\label{tab:metrics}
\end{table}

As was the case with the \emph{Datasets} and \emph{Methods} modules, by following the design principles upon which we built \emph{PhotoHolmes}, using one of the metrics included in \emph{PhotoHolmes} is a straightforward process. For example, to utilize the weighted IoU on its first version, the following code snippet can be employed:

\lstset{style=mystyle}

\begin{lstlisting}[language=Python]
from photoholmes.metrics import IoU_weighted_v1
import torch

iou_weighted_v1_metric = IoU_weighted_v1()

# Generate random data
data = [
    (torch.rand(256, 256), torch.randint(0, 2, (256, 256))) 
    for _ in range(10)
]

# Update the metric for each image
for pred, mask in data:
    iou_weighted_v1_metric.update(pred, mask)

# Compute the final value
iou_weighted = iou_weighted_v1_metric.compute()

print("IoU_weighted_v1:", iou_weighted)
\end{lstlisting}

The \emph{Metrics} module also contains a registry of the available metrics, as well as a factory that allows the user to load any of the registered metrics easily. Unlike the previous factories, the \emph{Metrics} factory can receive a list as input, returning the collection of metrics requested in one simple call.

\subsection{Benchmark}
\label{sec:benchmark}

This module is one of the most important features of the \textit{PhotoHolmes} library, and its importance is given by the ability to run evaluations easily and quickly while maintaining reproducibility and uniformity. The \emph{Benchmark} module, which consists of a single \emph{Benchmark} object, is designed to work seamlessly with any method, dataset, and metric that was implemented using \emph{PhotoHolmes}.

While one could build a custom benchmark script to run a method over a dataset, our module includes some useful functionalities that simplify the process. The most useful one is saving the outputs in a compressed NPZ format, making it possible to resume a benchmark process if it was interrupted or quickly re-evaluating the outputs on a different set of metrics without running the method again. It is important to note that the benchmark process can take a long time, especially if the method in place is slow and the dataset has a lot of images. Since batching requires the images to be of the same size, it is not possible to optimize inference speeds with this technique, as resizing and cropping can destroy useful traces in the image.

\begin{figure}[h]
    \centering
    \includegraphics[width = \columnwidth]{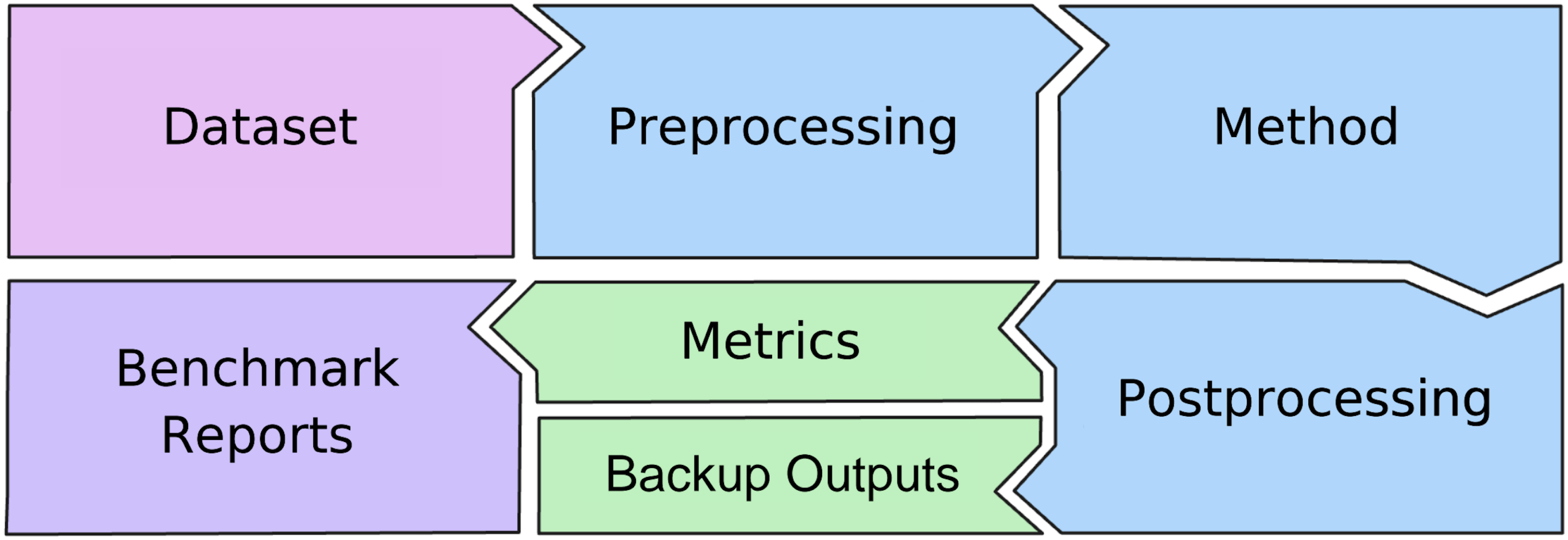}
    \caption{Benchmark class flow diagram. 
    Everything starts by choosing a dataset and a method, then according to the chosen method, the dataset is preprocessed with the corresponding preprocessing. Then, outputs can be visualized, and chosen metrics are computed. The metrics are then stored as benchmark reports. 
    }
    \label{C5:fig:benchmark_flow_diagram}
\end{figure}

The \emph{Benchmark} object is used in two steps. Firstly, we need to instantiate the class with the configurations the benchmark will follow. These configurations include the device to run on, whether to store outputs, re-use stored outputs and where to save them, and controlling the verbosity of the logging. Once we have our benchmark instance, we can call the \texttt{run} function, providing it a method, a dataset, and the set of metrics to run on. It is in this function that the \textit{benchmark} function required by the \emph{BaseMethod} is used. Figure~\ref{C5:fig:benchmark_flow_diagram} illustrates the described end-to-end benchmarking pipeline. 

During the state-of-the-art review, we identified three types of method outputs: heatmaps, binary masks, and detection scores. As such, we introduce this notion to our benchmark process, expecting a method to output at most one of each category. This way, we create an interface for the benchmark and the methods to interact. Each one of these output types is evaluated on a different set of metrics, 
resulting in a metric report for every output type the method under evaluation has.
A metric report is a JSON file, stored within the output folder, where the metric results for the method are dumped.

Once finished, the benchmark results will be in the output folder selected when creating the \emph{Benchmark} objects, which defaults to \texttt{output}. Inside this folder, the following structure is present:

\dirtree{%
    .1 output/.
    .2 \{method\}/.
    .3 \{dataset\}/.
    .4 metrics/.
    .5 \{timestamp\}\_\{dataset\_mode\}/.
    .6 \{output\_type\_1\}\_report.json. 
    .6 \{output\_type\_2\}\_report.json. 
    .4 outputs/.
    .5 \dots.
}

The following code snippet provides an example of using the \emph{Benchmark} module by concatenating all of the other modules included in \emph{PhotoHolmes}. The examples showcase the benchmarking of DQ in Columbia with AUROC and F1 by using the corresponding factories.

\lstset{style=mystyle}

\begin{lstlisting}[language=Python]
from photoholmes.datasets import DatasetFactory, DatasetRegistry
from photoholmes.metrics.factory import MetricFactory
from photoholmes.methods import MethodFactory, MethodRegistry
from photoholmes.benchmark import Benchmark

# Load the dataset
dataset = DatasetFactory.load(
    DatasetRegistry.COLUMBIA,
    dataset_path=columbia_dataset_path,
    load=["image", "dct_coefficients"],
    preprocessing_pipeline=dq_preprocessing,
)

# Load the metrics
metrics = MetricFactory.load(["auroc", "f1"])
print(metrics)

# Load the method
dq, dq_preprocessing = MethodFactory.load("dq")

# Create the Benchmark object
benchmark = Benchmark(
    save_method_outputs=True,
    save_extra_outputs=False,
    save_metrics=True,
    output_folder="example_output",
    device="cpu",
    use_existing_output=False,
    verbose=1,
)

# Run the benchmark
benchmark.run(method=dq, 
    dataset=dataset, 
    metrics=metrics
)
\end{lstlisting}

\subsection{Command Line Interface (CLI)}

As mentioned in the introduction, one of the design principles of the library is usability. Following this principle, a Command Line Interface (CLI) was developed to ease the user experience. In the first version of the library, the CLI contains three commands: \texttt{run}, \texttt{download\_weights}, \texttt{adapt\_weights}.

The \texttt{run} command allows the user to run a method in a single image and check the results without writing code. Each method has its sub-command and can expect more arguments (for instance, the path to the pre-trained weights in the case of learning-based methods), but they all share the following arguments and options:
\begin{itemize}
    \item Arguments
        \begin{itemize}
            \item \texttt{image\_path}: path to the image to run the method on.
        \end{itemize}
    \item Options
    \begin{itemize}
        \item \texttt{output-folder}: path to a folder where to save the method outputs. If no path
                     is provided, then the outputs are not saved.
        \item \texttt{overlay}: flag that, if set, a plot with the mask or heatmap overlayed on the image is included.
        \item \texttt{show-plot} / \texttt{no-show-plot}: whether to show results as a matplotlib plot.
        \item \texttt{device}: torch device to run the methods on. Only available in methods that use neural networks.
    \end{itemize}
\end{itemize}

\begin{figure}
    \centering
    \includegraphics[width=\textwidth]{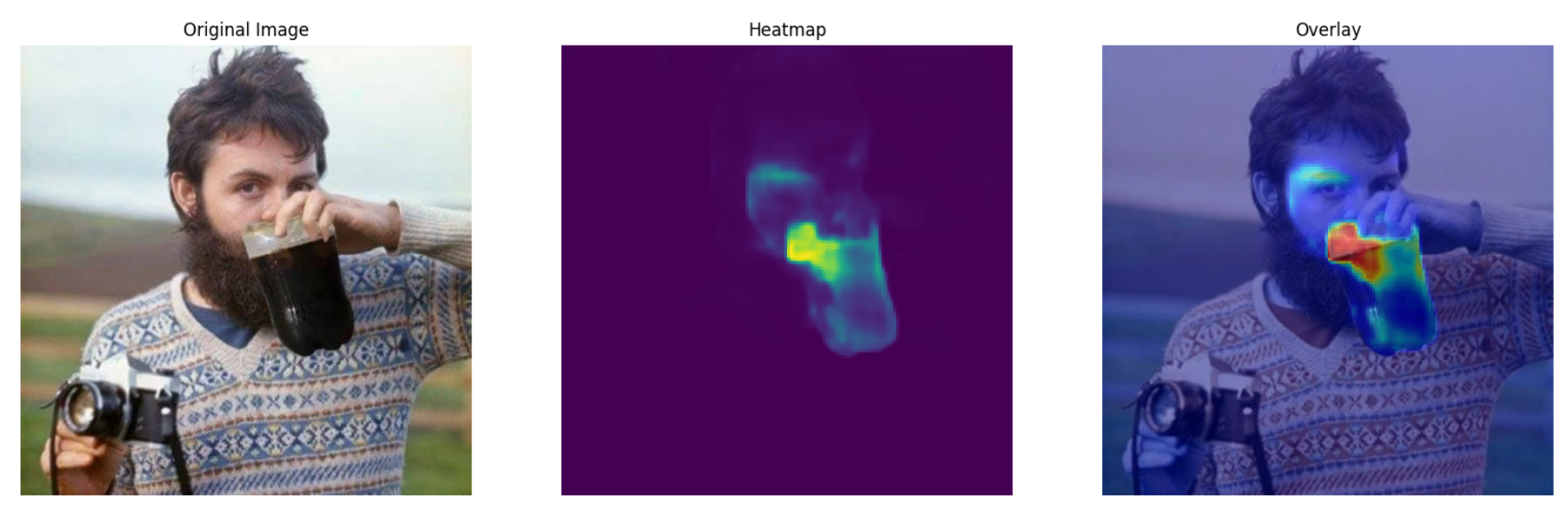}
    \caption{Output of running \texttt{photoholmes run catnet <image\_path> --overlay} using the  CLI. The forged image is the one presented in 
    Figure~\ref{fig:outputs}.}
    \label{C5:fig:cli_res}
\end{figure}

In Figure~\ref{C5:fig:cli_res}, we present the output of running CAT-Net on the forged image from Figure~\ref{fig:outputs}. To get that result, the user can run \texttt{photoholmes run catnet <image\_path> --overlay} using the CLI.

The \texttt{download\_weights} command provides a simple interface for users to download the model's weights, in case of deep learning methods. This command takes a method as an argument and has the option to choose the folder where the weights are downloaded. As mentioned before, some of the methods included in \mbox{\emph{PhotoHolmes}} have their weights licensed to be used only in research contexts. When this is the case, the CLI will display a warning and require the user's input on whether they accept those terms or not, protecting the original author's rights.

Lastly, the \texttt{adapt\_weights} script modifies original model weights to align with \emph{PhotoHolmes'} methods implementations. Some methods have been adjusted to eliminate unnecessary structures within the architecture and remnants from applying transfer learning. This streamlines the model, making it more efficient and suitable for the intended tasks. For example, EXIF as Language~\cite{zheng2023exif} inherits from \emph{OpenAI}'s Clip model~\cite{radford2021learning}, yet overrides some of its properties and leaves some structures unused. Our scripts remove any unused modules for a cleaner implementation. 

\subsection{Contributing to \textit{PhotoHolmes}}

As it was mentioned in Section~\ref{sec:design_principles}, \textit{PhotoHolmes} was designed with extensibility in mind.
This means users are invited to contribute to the library, by integrating methods, datasets or metrics of interest. This can be done easily by extending the basic structures we provide in each module.
Note that each of these sections has its own README file where the steps for doing so are thoroughly specified.

\section{Use cases}
\label{sec:use_cases}

Figure~\ref{fig:outputs} shows the results of running the \textit{PhotoHolmes} CLI for all methods on a  given suspicious image. To do this, the user has to use the \texttt{run} command for the different methods. The user has the choice to save the outputs or to just preview them in a pop-up figure. This illustrates how \textit{PhotoHolmes} can be used to easily check for forged areas on suspicious images, and given the wide array of included methods, the chances of getting a good localization of the forgery are maximized. 

Another interesting use case can be seen when using the previously mentioned \textit{Benchmark} module. A code similar to that exposed in the Benchmark section (Section~\ref{sec:benchmark}) can yield results for evaluating a method on a dataset, with a set of metrics. These results are saved into a JSON file which can be later processed for further analysis. By iterating over this script with different datasets and methods, users can develop tables of results such as Tables~\ref{tab:localization_trace_f1_v1} and~\ref{tab:localization_f1_popular_v1} that show the mean weighted F1 score (F1$^{v_1}_w$) for localization in all datasets and Table~\ref{tab:detection_popular_F1} that shows the weighted dataset-level F1 score (F1$^{v_2}_w$) for detection in all datasets that have both pristine and forged images. 

We will not delve into these results in particular, as it is not within the scope of this article, however an in-depth analysis of results can be found in~\cite{tesis_photoholmes}. Yet, it is clear that such results allow for an interesting analysis, and it is through \textit{PhotoHolmes} that one can obtain them with little code development, and in a reproducible manner. This, added to the fact that the library's \texttt{datasets} and \texttt{methods} modules can be easily extended with new custom datasets and methods, is what we believe can make \texttt{PhotoHolmes} of great contribution to the image forensics community.

\begin{table}[]
\centering
\resizebox{\textwidth}{!}{%
\begin{tabular}{lcccccc}
\toprule
Method                                                             & Noise                                                                        & JPEG Quality                                          & JPEG Grid                                             & CFA Alg.                                               & CFA Grid                                              & Hybrid                                                \\ 
\midrule
& & & & & & \\[-8
pt]
Adaptive CFA                                                       & {\begin{tabular}[c]{@{}c@{}}\textcolor{blue}{0.023}\\ \textcolor{codegreen}{0.039}\end{tabular}} & \begin{tabular}[c]{@{}c@{}}\textcolor{blue}{0.197}\\ \textcolor{codegreen}{0.191}\end{tabular} & \begin{tabular}[c]{@{}c@{}}\textcolor{blue}{0.198}\\ \textcolor{codegreen}{0.197}\end{tabular} & \begin{tabular}[c]{@{}c@{}}\textcolor{blue}{\textbf{0.531}}\\ \textcolor{codegreen}{\textbf{0.531}}\end{tabular} & \begin{tabular}[c]{@{}c@{}}\textcolor{blue}{\textbf{0.692}}\\ \textcolor{codegreen}{\textbf{0.676}}\end{tabular} & \begin{tabular}[c]{@{}c@{}}\textcolor{blue}{0.265}\\ \textcolor{codegreen}{0.260}\end{tabular} \\[9pt] 
Noisesniffer                                                       & \begin{tabular}[c]{@{}c@{}}\textcolor{blue}{\textbf{0.255}}\\ \textcolor{codegreen}{\underline{0.199}}\end{tabular}        & \begin{tabular}[c]{@{}c@{}}\textcolor{blue}{0.168}\\ \textcolor{codegreen}{0.126}\end{tabular} & \begin{tabular}[c]{@{}c@{}}\textcolor{blue}{0.075}\\ \textcolor{codegreen}{0.082}\end{tabular} & \begin{tabular}[c]{@{}c@{}}\textcolor{blue}{0.154}\\ \textcolor{codegreen}{0.116}\end{tabular} & \begin{tabular}[c]{@{}c@{}}\textcolor{blue}{0.072}\\ \textcolor{codegreen}{0.070}\end{tabular} & \begin{tabular}[c]{@{}c@{}}\textcolor{blue}{0.222}\\ \textcolor{codegreen}{0.173}\end{tabular} \\[9pt] 
ZERO                                                               & \begin{tabular}[c]{@{}c@{}}\textcolor{blue}{0.000}\\ \textcolor{codegreen}{0.000}\end{tabular} & \begin{tabular}[c]{@{}c@{}}\textcolor{blue}{\textbf{0.747}}\\ \textcolor{codegreen}{\textbf{0.683}}\end{tabular} & \begin{tabular}[c]{@{}c@{}}\textcolor{blue}{\textbf{0.785}}\\ \textcolor{codegreen}{\textbf{0.697}}\end{tabular} & \begin{tabular}[c]{@{}c@{}}\textcolor{blue}{0.000}\\ \textcolor{codegreen}{0.000}\end{tabular} & \begin{tabular}[c]{@{}c@{}}\textcolor{blue}{0.000}\\ \textcolor{codegreen}{0.000}\end{tabular} & \begin{tabular}[c]{@{}c@{}}\textcolor{blue}{\textbf{0.608}}\\ \textcolor{codegreen}{\textbf{0.604}}\end{tabular} \\[9pt]
\begin{tabular}[c]{@{}c@{}}ZERO with\\ missing grids\end{tabular}                                                              & \begin{tabular}[c]{@{}c@{}}\textcolor{blue}{0.000}\\ \textcolor{codegreen}{0.000}\end{tabular} & \begin{tabular}[c]{@{}c@{}}\textcolor{blue}{\underline{0.699}}\\ \textcolor{codegreen}{\underline{0.660}}\end{tabular} & \begin{tabular}[c]{@{}c@{}}\textcolor{blue}{\underline{0.711}}\\ \textcolor{codegreen}{\underline{0.662}}\end{tabular} & \begin{tabular}[c]{@{}c@{}}\textcolor{blue}{0.000}\\ \textcolor{codegreen}{0.000}\end{tabular} & \begin{tabular}[c]{@{}c@{}}\textcolor{blue}{0.000}\\ \textcolor{codegreen}{0.000}\end{tabular} & \begin{tabular}[c]{@{}c@{}}\textcolor{blue}{\underline{0.576}}\\ \textcolor{codegreen}{\underline{0.567}}\end{tabular} \\[9pt]
DQ                                                                 & \begin{tabular}[c]{@{}c@{}}\textcolor{blue}{0.163}\\ \textcolor{codegreen}{0.163}\end{tabular} & \begin{tabular}[c]{@{}c@{}}\textcolor{blue}{0.207}\\ \textcolor{codegreen}{0.202}\end{tabular} & \begin{tabular}[c]{@{}c@{}}\textcolor{blue}{0.209}\\ \textcolor{codegreen}{0.204}\end{tabular} & \begin{tabular}[c]{@{}c@{}}\textcolor{blue}{0.170}\\ \textcolor{codegreen}{0.169}\end{tabular} & \begin{tabular}[c]{@{}c@{}}\textcolor{blue}{0.169}\\ \textcolor{codegreen}{0.169}\end{tabular} & \begin{tabular}[c]{@{}c@{}}\textcolor{blue}{0.202}\\ \textcolor{codegreen}{0.199}\end{tabular} \\[9pt] 
CAT-Net                                                             & \begin{tabular}[c]{@{}c@{}}\textcolor{blue}{0.009}\\ \textcolor{codegreen}{0.006}\end{tabular}                        & \begin{tabular}[c]{@{}c@{}}\textcolor{blue}{0.276}\\ \textcolor{codegreen}{0.319}\end{tabular} & \begin{tabular}[c]{@{}c@{}}\textcolor{blue}{0.270}\\ \textcolor{codegreen}{0.319}\end{tabular} & \begin{tabular}[c]{@{}c@{}}\textcolor{blue}{0.005}\\ \textcolor{codegreen}{0.007}\end{tabular} & \begin{tabular}[c]{@{}c@{}}\textcolor{blue}{0.005}\\ \textcolor{codegreen}{0.004}\end{tabular} & \begin{tabular}[c]{@{}c@{}}\textcolor{blue}{0.327}\\ \textcolor{codegreen}{0.350}\end{tabular} \\[9pt] 
Splicebuster                                                            & \begin{tabular}[c]{@{}c@{}}\textcolor{blue}{0.116}\\ \textcolor{codegreen}{0.118}\end{tabular} & \begin{tabular}[c]{@{}c@{}}\textcolor{blue}{0.124}\\ \textcolor{codegreen}{0.132}\end{tabular} & \begin{tabular}[c]{@{}c@{}}\textcolor{blue}{0.070}\\ \textcolor{codegreen}{0.088}\end{tabular} & \begin{tabular}[c]{@{}c@{}}\textcolor{blue}{0.096}\\ \textcolor{codegreen}{0.124}\end{tabular} & \begin{tabular}[c]{@{}c@{}}\textcolor{blue}{0.060}\\ \textcolor{codegreen}{0.077}\end{tabular} & \begin{tabular}[c]{@{}c@{}}\textcolor{blue}{0.147}\\ \textcolor{codegreen}{0.150}\end{tabular} \\[9pt] 
\begin{tabular}[l]{@{}l@{}}EXIF\\ Mean Shift\end{tabular} & \begin{tabular}[c]{@{}c@{}}\textcolor{blue}{0.101}\\ \textcolor{codegreen}{0.111}\end{tabular} & \begin{tabular}[c]{@{}c@{}}\textcolor{blue}{0.146}\\ \textcolor{codegreen}{0.145}\end{tabular} & \begin{tabular}[c]{@{}c@{}}\textcolor{blue}{0.090}\\ \textcolor{codegreen}{0.103}\end{tabular} & \begin{tabular}[c]{@{}c@{}}\textcolor{blue}{0.159}\\ \textcolor{codegreen}{0.165}\end{tabular} & \begin{tabular}[c]{@{}c@{}}\textcolor{blue}{0.110}\\ \textcolor{codegreen}{0.127}\end{tabular} & \begin{tabular}[c]{@{}c@{}}\textcolor{blue}{0.123}\\ \textcolor{codegreen}{0.136}\end{tabular} \\[9pt]
\begin{tabular}[l]{@{}c@{}}EXIF\\ Ncuts\end{tabular}    & \begin{tabular}[c]{@{}c@{}}\textcolor{blue}{0.189}\\ \textcolor{codegreen}{\textbf{0.225}}\end{tabular} & \begin{tabular}[c]{@{}c@{}}\textcolor{blue}{0.221}\\ \textcolor{codegreen}{0.240}\end{tabular} & \begin{tabular}[c]{@{}c@{}}\textcolor{blue}{0.176}\\ \textcolor{codegreen}{0.212}\end{tabular} & \begin{tabular}[c]{@{}c@{}}\textcolor{blue}{\underline{0.227}}\\ \textcolor{codegreen}{0.243}\end{tabular} & \begin{tabular}[c]{@{}c@{}}\textcolor{blue}{0.160}\\ \textcolor{codegreen}{\underline{0.193}}\end{tabular} & \begin{tabular}[c]{@{}c@{}}\textcolor{blue}{0.234}\\ \textcolor{codegreen}{0.284}\end{tabular} \\[9pt] 
PSCC-Net                                                            & \begin{tabular}[c]{@{}c@{}}\textcolor{blue}{0.189}\\ \textcolor{codegreen}{0.188}\end{tabular} & \begin{tabular}[c]{@{}c@{}}\textcolor{blue}{0.182}\\ \textcolor{codegreen}{0.176}\end{tabular} & \begin{tabular}[c]{@{}c@{}}\textcolor{blue}{0.170}\\ \textcolor{codegreen}{0.173}\end{tabular} & \begin{tabular}[c]{@{}c@{}}\textcolor{blue}{0.198}\\ \textcolor{codegreen}{\underline{0.254}}\end{tabular} & \begin{tabular}[c]{@{}c@{}}\textcolor{blue}{\underline{0.196}}\\ \textcolor{codegreen}{\underline{0.193}}\end{tabular} & \begin{tabular}[c]{@{}c@{}}\textcolor{blue}{0.189}\\ \textcolor{codegreen}{0.183}\end{tabular} \\[9pt] 
TruFor                                                             & \begin{tabular}[c]{@{}c@{}}\textcolor{blue}{\underline{0.198}}\\ \textcolor{codegreen}{0.145}\end{tabular} & \begin{tabular}[c]{@{}c@{}}\textcolor{blue}{0.542}\\ \textcolor{codegreen}{0.561}\end{tabular} & \begin{tabular}[c]{@{}c@{}}\textcolor{blue}{0.564}\\ \textcolor{codegreen}{0.583}\end{tabular} & \begin{tabular}[c]{@{}c@{}}\textcolor{blue}{0.114}\\ \textcolor{codegreen}{0.103}\end{tabular} & \begin{tabular}[c]{@{}c@{}}\textcolor{blue}{0.077}\\ \textcolor{codegreen}{0.077}\end{tabular} & \begin{tabular}[c]{@{}c@{}}\textcolor{blue}{0.458}\\ \textcolor{codegreen}{0.446}\end{tabular} \\[9pt]
FOCAL                                                             & \begin{tabular}[c]{@{}c@{}}\textcolor{blue}{0.149}\\ \textcolor{codegreen}{0.151}\end{tabular} & \begin{tabular}[c]{@{}c@{}}\textcolor{blue}{0.118}\\ \textcolor{codegreen}{0.137}\end{tabular} & \begin{tabular}[c]{@{}c@{}}\textcolor{blue}{0.116}\\ \textcolor{codegreen}{0.135}\end{tabular} & \begin{tabular}[c]{@{}c@{}}\textcolor{blue}{0.117}\\ \textcolor{codegreen}{0.131}\end{tabular} & \begin{tabular}[c]{@{}c@{}}\textcolor{blue}{0.122}\\ \textcolor{codegreen}{0.129}\end{tabular} & \begin{tabular}[c]{@{}c@{}}\textcolor{blue}{0.146}\\ \textcolor{codegreen}{0.163}\end{tabular} \\ 
\bottomrule
\end{tabular}%
}
\caption{Localization performance in terms of the mean weighted F1 score (F1$^{v_1}_w$) in the Trace database, for both, the \textcolor{blue}{exogenous datasets} and the \textcolor{codegreen}{endogenous datasets}. In \textbf{bold}, the highest score in each dataset, and \underline{underlined}, the second highest one.}
\label{tab:localization_trace_f1_v1}
\end{table}
\begin{table}[]
\centering
\setlength{\tabcolsep}{2pt}
\resizebox{\columnwidth}{!}{%
\begin{tabular}{lccccccccccc}
\toprule
Method                                                             & Columbia                                                      & \begin{tabular}[c]{@{}c@{}}Columbia\\ WebP\end{tabular}   & \begin{tabular}[c]{@{}c@{}}Casia1.0\\ SP\end{tabular}         & \begin{tabular}[c]{@{}c@{}}Casia1.0\\ CM\end{tabular}         & COVERAGE                                                  & DSO-1                                                         & Korus                                                     & \begin{tabular}[c]{@{}c@{}}Korus\\ WebP\end{tabular}      & \begin{tabular}[c]{@{}c@{}}AutoSplice\\ 100\end{tabular}  & \begin{tabular}[c]{@{}c@{}}AutoSplice\\ 90\end{tabular} & \begin{tabular}[c]{@{}c@{}}AutoSplice\\ 75\end{tabular} \\[8pt] \midrule
\begin{tabular}[l]{@{}l@{}}Adaptive \\ CFA \\ \end{tabular} & 
 \begin{tabular}[c]{@{}c@{}}\textcolor{blue}{0.183}\\ \textcolor{codegreen}{0.370}\\ \textcolor{codepurple}{0.314}\end{tabular} & \begin{tabular}[c]{@{}c@{}}\textcolor{blue}{0.181}\\ \textcolor{codegreen}{0.366}\\ \textcolor{codepurple}{-}\end{tabular} & \begin{tabular}[c]{@{}c@{}}\textcolor{blue}{0.066}\\ \textcolor{codegreen}{0.178}\\ \textcolor{codepurple}{0.171}\end{tabular} & \begin{tabular}[c]{@{}c@{}}\textcolor{blue}{0.048}\\ \textcolor{codegreen}{0.129}\\ \textcolor{codepurple}{0.129}\end{tabular} & \begin{tabular}[c]{@{}c@{}}\textcolor{blue}{0.096}\\ \textcolor{codegreen}{0.201}\\ \textcolor{codepurple}{-}\end{tabular} & \begin{tabular}[c]{@{}c@{}}\textcolor{blue}{0.120}\\ \textcolor{codegreen}{0.246}\\ \textcolor{codepurple}{0.202}\end{tabular} & \begin{tabular}[c]{@{}c@{}}\textcolor{blue}{0.102}\\ \textcolor{codegreen}{0.204}\\ \textcolor{codepurple}{-}\end{tabular} & \begin{tabular}[c]{@{}c@{}}\textcolor{blue}{0.049}\\ \textcolor{codegreen}{0.099}\\ \textcolor{codepurple}{-}\end{tabular} & \begin{tabular}[c]{@{}c@{}}\textcolor{blue}{0.239}\\ \textcolor{codegreen}{0.389}\\ \textcolor{codepurple}{-}\end{tabular} & \begin{tabular}[c]{@{}c@{}}\textcolor{blue}{-}\\ \textcolor{codegreen}{0.379}\\ \textcolor{codepurple}{-}\end{tabular}   & \begin{tabular}[c]{@{}c@{}}\textcolor{blue}{-}\\ \textcolor{codegreen}{0.375}\\ \textcolor{codepurple}{-}\end{tabular}   \\[22pt]
Noisesniffer                                                       & \begin{tabular}[c]{@{}c@{}}\textcolor{blue}{0.050}\\ \textcolor{codegreen}{0.101}\\ \textcolor{codepurple}{0.044}\end{tabular} & \begin{tabular}[c]{@{}c@{}}\textcolor{blue}{0.012}\\ \textcolor{codegreen}{0.023}\\ \textcolor{codepurple}{-}\end{tabular} & \begin{tabular}[c]{@{}c@{}}\textcolor{blue}{0.027}\\ \textcolor{codegreen}{0.073}\\ \textcolor{codepurple}{0.074}\end{tabular} & \begin{tabular}[c]{@{}c@{}}\textcolor{blue}{0.013}\\ \textcolor{codegreen}{0.036}\\ \textcolor{codepurple}{0.036}\end{tabular} & \begin{tabular}[c]{@{}c@{}}\textcolor{blue}{0.029} \\ \textcolor{codegreen}{0.062}\\ \textcolor{codepurple}{-}\end{tabular} & \begin{tabular}[c]{@{}c@{}}\textcolor{blue}{0.139}\\ \textcolor{codegreen}{0.285}\\ \textcolor{codepurple}{0.273}\end{tabular} & \begin{tabular}[c]{@{}c@{}}\textcolor{blue}{0.099}\\ \textcolor{codegreen}{0.197}\\ \textcolor{codepurple}{-}\end{tabular} & \begin{tabular}[c]{@{}c@{}}\textcolor{blue}{0.065}\\ \textcolor{codegreen}{0.130}\\ \textcolor{codepurple}{-}\end{tabular} & \begin{tabular}[c]{@{}c@{}}\textcolor{blue}{0.033}\\ \textcolor{codegreen}{0.053}\\ \textcolor{codepurple}{-}\end{tabular} & \begin{tabular}[c]{@{}c@{}}\textcolor{blue}{-}\\ \textcolor{codegreen}{0.038}\\ \textcolor{codepurple}{-}\end{tabular}   & \begin{tabular}[c]{@{}c@{}}\textcolor{blue}{-}\\ \textcolor{codegreen}{0.041}\\ \textcolor{codepurple}{-}\end{tabular}   \\[22pt]
ZERO                                                               & \begin{tabular}[c]{@{}c@{}}\textcolor{blue}{0.000}\\ \textcolor{codegreen}{0.000}\\ \textcolor{codepurple}{0.000}\end{tabular} & \begin{tabular}[c]{@{}c@{}}\textcolor{blue}{0.006}\\ \textcolor{codegreen}{0.011}\\ \textcolor{codepurple}{-}\end{tabular} & \begin{tabular}[c]{@{}c@{}}\textcolor{blue}{0.020}\\ \textcolor{codegreen}{0.054}\\ \textcolor{codepurple}{0.000}\end{tabular} & \begin{tabular}[c]{@{}c@{}}\textcolor{blue}{0.000}\\ \textcolor{codegreen}{0.001}\\ \textcolor{codepurple}{0.000}\end{tabular} & \begin{tabular}[c]{@{}c@{}}\textcolor{blue}{0.004}\\ \textcolor{codegreen}{0.009}\\ \textcolor{codepurple}{-}\end{tabular} & \begin{tabular}[c]{@{}c@{}}\textcolor{blue}{0.057}\\ \textcolor{codegreen}{0.116}\\ \textcolor{codepurple}{0.000}\end{tabular} & \begin{tabular}[c]{@{}c@{}}\textcolor{blue}{0.002}\\ \textcolor{codegreen}{0.004}\\ \textcolor{codepurple}{-}\end{tabular} & \begin{tabular}[c]{@{}c@{}}\textcolor{blue}{0.021}\\ \textcolor{codegreen}{0.042}\\ \textcolor{codepurple}{-}\end{tabular} & \begin{tabular}[c]{@{}c@{}}\textcolor{blue}{0.000}\\ \textcolor{codegreen}{0.000}\\ \textcolor{codepurple}{-}\end{tabular} & \begin{tabular}[c]{@{}c@{}}\textcolor{blue}{-}\\ \textcolor{codegreen}{0.000}\\ \textcolor{codepurple}{-}\end{tabular}   & \begin{tabular}[c]{@{}c@{}}\textcolor{blue}{-}\\ \textcolor{codegreen}{0.000}\\ \textcolor{codepurple}{-}\end{tabular}   \\[22pt]

\begin{tabular}[l]{@{}l@{}}ZERO with\\ missing grids\end{tabular}                         & 
\begin{tabular}[c]{@{}c@{}}\textcolor{blue}{0.002}\\ \textcolor{codegreen}{0.002}\\ \textcolor{codepurple}{0.001}\end{tabular} & \begin{tabular}[c]{@{}c@{}}\textcolor{blue}{0.053}\\ \textcolor{codegreen}{0.107}\\ \textcolor{codepurple}{-}\end{tabular} & \begin{tabular}[c]{@{}c@{}}\textcolor{blue}{0.021}\\ \textcolor{codegreen}{0.057}\\ \textcolor{codepurple}{0.003}\end{tabular} & \begin{tabular}[c]{@{}c@{}}\textcolor{blue}{0.007}\\ \textcolor{codegreen}{0.019}\\ \textcolor{codepurple}{0.020}\end{tabular} & \begin{tabular}[c]{@{}c@{}}\textcolor{blue}{0.006}\\ \textcolor{codegreen}{0.013}\\ \textcolor{codepurple}{-}\end{tabular} & \begin{tabular}[c]{@{}c@{}}\textcolor{blue}{0.158}\\ \textcolor{codegreen}{0.324}\\ \textcolor{codepurple}{0.008}\end{tabular} & \begin{tabular}[c]{@{}c@{}}\textcolor{blue}{0.003}\\ \textcolor{codegreen}{0.006}\\ \textcolor{codepurple}{-}\end{tabular} & \begin{tabular}[c]{@{}c@{}}\textcolor{blue}{0.039}\\ \textcolor{codegreen}{0.094}\\ \textcolor{codepurple}{-}\end{tabular} & \begin{tabular}[c]{@{}c@{}}\textcolor{blue}{0.236}\\ \textcolor{codegreen}{0.384}\\ \textcolor{codepurple}{-}\end{tabular} & \begin{tabular}[c]{@{}c@{}}\textcolor{blue}{-}\\ \textcolor{codegreen}{0.008}\\ \textcolor{codepurple}{-}\end{tabular}   & \begin{tabular}[c]{@{}c@{}}\textcolor{blue}{-}\\ \textcolor{codegreen}{0.010}\\ \textcolor{codepurple}{-}\end{tabular}   \\[22pt]
DQ                                                                 & \begin{tabular}[c]{@{}c@{}}\textcolor{blue}{0.148}\\ \textcolor{codegreen}{0.299}\\ \textcolor{codepurple}{0.211}\end{tabular} & \begin{tabular}[c]{@{}c@{}}\textcolor{blue}{0.151}\\ \textcolor{codegreen}{0.305}\\ \textcolor{codepurple}{-}\end{tabular} & \begin{tabular}[c]{@{}c@{}}\textcolor{blue}{0.038}\\ \textcolor{codegreen}{0.104}\\ \textcolor{codepurple}{0.099}\end{tabular} & \begin{tabular}[c]{@{}c@{}}\textcolor{blue}{0.025}\\ \textcolor{codegreen}{0.068}\\ \textcolor{codepurple}{0.069}\end{tabular} & \begin{tabular}[c]{@{}c@{}}\textcolor{blue}{0.083}\\ \textcolor{codegreen}{0.175}\\ \textcolor{codepurple}{-}\end{tabular} & \begin{tabular}[c]{@{}c@{}}\textcolor{blue}{0.094}\\ \textcolor{codegreen}{0.193}\\ \textcolor{codepurple}{0.128}\end{tabular} & \begin{tabular}[c]{@{}c@{}}\textcolor{blue}{0.043}\\ \textcolor{codegreen}{0.087}\\ \textcolor{codepurple}{-}\end{tabular} & \begin{tabular}[c]{@{}c@{}}\textcolor{blue}{0.044}\\ \textcolor{codegreen}{0.087}\\ \textcolor{codepurple}{-}\end{tabular} & \begin{tabular}[c]{@{}c@{}}\textcolor{blue}{0.241}\\ \textcolor{codegreen}{0.393}\\ \textcolor{codepurple}{-}\end{tabular} & \begin{tabular}[c]{@{}c@{}}\textcolor{blue}{-}\\ \textcolor{codegreen}{0.268}\\ \textcolor{codepurple}{-}\end{tabular}   & \begin{tabular}[c]{@{}c@{}}\textcolor{blue}{-}\\ \textcolor{codegreen}{0.154}\\ \textcolor{codepurple}{-}\end{tabular}   \\[22pt]
CAT-Net                                                             &
\begin{tabular}[c]{@{}c@{}}\textcolor{blue}{\underline{0.383}}\\ \textcolor{codegreen}{\underline{0.773}}\\ \textcolor{codepurple}{\underline{0.902}}\end{tabular} &
\begin{tabular}[c]{@{}c@{}}\textcolor{blue}{0.353}\\ \textcolor{codegreen}{0.713}\\ \textcolor{codepurple}{-}\end{tabular} &
\begin{tabular}[c]{@{}c@{}}\textcolor{blue}{\underline{0.271}}\\ \textcolor{codegreen}{\underline{0.735}}\\ \textcolor{codepurple}{0.589}\end{tabular} &
\begin{tabular}[c]{@{}c@{}}\textcolor{blue}{\underline{0.212}}\\ \textcolor{codegreen}{\underline{0.577}}\\ \textcolor{codepurple}{\underline{0.563}}\end{tabular} &
\begin{tabular}[c]{@{}c@{}}\textcolor{blue}{0.154}\\ \textcolor{codegreen}{0.322}\\ \textcolor{codepurple}{-}\end{tabular} &
\begin{tabular}[c]{@{}c@{}}\textcolor{blue}{0.231}\\ \textcolor{codegreen}{0.474}\\ \textcolor{codepurple}{0.125}\end{tabular} &
\begin{tabular}[c]{@{}c@{}}\textcolor{blue}{0.033}\\ \textcolor{codegreen}{0.066}\\ \textcolor{codepurple}{-}\end{tabular} &
\begin{tabular}[c]{@{}c@{}}\textcolor{blue}{0.027}\\ \textcolor{codegreen}{0.053}\\ \textcolor{codepurple}{-}\end{tabular} &
\begin{tabular}[c]{@{}c@{}}\textcolor{blue}{\textbf{0.513}}\\ \textcolor{codegreen}{\textbf{0.835}}\\ \textcolor{codepurple}{-}\end{tabular} &
\begin{tabular}[c]{@{}c@{}}\textcolor{blue}{-}\\ \textcolor{codegreen}{\textbf{0.767}}\\ \textcolor{codepurple}{-}\end{tabular}   &
\begin{tabular}[c]{@{}c@{}}\textcolor{blue}{-}\\ \textcolor{codegreen}{\textbf{0.433}}\\ \textcolor{codepurple}{-}\end{tabular}   \\[22pt]
Splicebuster                                                       & \begin{tabular}[c]{@{}c@{}}\textcolor{blue}{0.203}\\ \textcolor{codegreen}{0.410}\\ \textcolor{codepurple}{0.260}\end{tabular} & \begin{tabular}[c]{@{}c@{}}\textcolor{blue}{0.091}\\ \textcolor{codegreen}{0.183}\\ \textcolor{codepurple}{-}\end{tabular} & \begin{tabular}[c]{@{}c@{}}\textcolor{blue}{0.027}\\ \textcolor{codegreen}{0.075}\\ \textcolor{codepurple}{0.071}\end{tabular} & \begin{tabular}[c]{@{}c@{}}\textcolor{blue}{0.022}\\ \textcolor{codegreen}{0.061}\\ \textcolor{codepurple}{0.063}\end{tabular} & \begin{tabular}[c]{@{}c@{}}\textcolor{blue}{0.040}\\ \textcolor{codegreen}{0.085}\\ \textcolor{codepurple}{-}\end{tabular} & \begin{tabular}[c]{@{}c@{}}\textcolor{blue}{0.150}\\ \textcolor{codegreen}{0.307}\\ \textcolor{codepurple}{0.196}\end{tabular} & \begin{tabular}[c]{@{}c@{}}\textcolor{blue}{0.084}\\ \textcolor{codegreen}{0.168}\\ \textcolor{codepurple}{-}\end{tabular} & \begin{tabular}[c]{@{}c@{}}\textcolor{blue}{0.052}\\ \textcolor{codegreen}{0.103}\\ \textcolor{codepurple}{-}\end{tabular}     & \begin{tabular}[c]{@{}c@{}}\textcolor{blue}{0.086}\\ \textcolor{codegreen}{0.140}\\ \textcolor{codepurple}{-}\end{tabular} & \begin{tabular}[c]{@{}c@{}}\textcolor{blue}{-}\\ \textcolor{codegreen}{0.150}\\ \textcolor{codepurple}{-}\end{tabular}   & \begin{tabular}[c]{@{}c@{}}\textcolor{blue}{-}\\ \textcolor{codegreen}{0.152}\\ \textcolor{codepurple}{-}\end{tabular}   \\[22pt]
\begin{tabular}[l]{@{}l@{}}EXIF\\ Mean Shift\\\end{tabular} & \begin{tabular}[c]{@{}c@{}}\textcolor{blue}{0.227}\\ \textcolor{codegreen}{0.458}\\ \textcolor{codepurple}{0.356}\end{tabular} & \begin{tabular}[c]{@{}c@{}}\textcolor{blue}{0.135}\\ \textcolor{codegreen}{0.274}\\ \textcolor{codepurple}{-}\end{tabular}         & \begin{tabular}[c]{@{}c@{}}\textcolor{blue}{0.034}\\ \textcolor{codegreen}{0.093}\\ \textcolor{codepurple}{0.087}\end{tabular} & \begin{tabular}[c]{@{}c@{}}\textcolor{blue}{0.019}\\ \textcolor{codegreen}{0.052}\\ \textcolor{codepurple}{0.053}\end{tabular} & \begin{tabular}[c]{@{}c@{}}\textcolor{blue}{0.062}\\ \textcolor{codegreen}{0.130}\\ \textcolor{codepurple}{-}\end{tabular} & \begin{tabular}[c]{@{}c@{}}\textcolor{blue}{0.167}\\ \textcolor{codegreen}{0.344}\\ \textcolor{codepurple}{0.253}\end{tabular} & \begin{tabular}[c]{@{}c@{}}\textcolor{blue}{0.048}\\ \textcolor{codegreen}{0.096}\\ \textcolor{codepurple}{-}\end{tabular} & \begin{tabular}[c]{@{}c@{}}\textcolor{blue}{0.040}\\ \textcolor{codegreen}{0.080}\\ \textcolor{codepurple}{-}\end{tabular} & \begin{tabular}[c]{@{}c@{}}\textcolor{blue}{0.124}\\ \textcolor{codegreen}{0.201}\\ \textcolor{codepurple}{-}\end{tabular} & \begin{tabular}[c]{@{}c@{}}\textcolor{blue}{-}\\ \textcolor{codegreen}{0.150}\\ \textcolor{codepurple}{-}\end{tabular}   & \begin{tabular}[c]{@{}c@{}}\textcolor{blue}{-}\\ \textcolor{codegreen}{0.124}\\ \textcolor{codepurple}{-}\end{tabular}   \\[22pt]
\begin{tabular}[l]{@{}l@{}}EXIF\\ NCuts\\ \end{tabular}    & \begin{tabular}[c]{@{}c@{}}\textcolor{blue}{0.302}\\ \textcolor{codegreen}{0.609}\\ \textcolor{codepurple}{0.495}\end{tabular} & \begin{tabular}[c]{@{}c@{}}\textcolor{blue}{0.210}\\ \textcolor{codegreen}{0.425}\\ \textcolor{codepurple}{-}\end{tabular} & \begin{tabular}[c]{@{}c@{}}\textcolor{blue}{0.068}\\ \textcolor{codegreen}{0.185}\\ \textcolor{codepurple}{0.158}\end{tabular} & \begin{tabular}[c]{@{}c@{}}\textcolor{blue}{0.037}\\ \textcolor{codegreen}{0.102}\\ \textcolor{codepurple}{0.103}\end{tabular} & \begin{tabular}[c]{@{}c@{}}\textcolor{blue}{0.064}\\ \textcolor{codegreen}{0.133}\\ \textcolor{codepurple}{-}\end{tabular} & \begin{tabular}[c]{@{}c@{}}\textcolor{blue}{0.213}\\ \textcolor{codegreen}{0.437}\\ \textcolor{codepurple}{0.387}\end{tabular} & \begin{tabular}[c]{@{}c@{}}\textcolor{blue}{0.062}\\ \textcolor{codegreen}{0.124}\\ \textcolor{codepurple}{-}\end{tabular} & \begin{tabular}[c]{@{}c@{}}\textcolor{blue}{0.050}\\ \textcolor{codegreen}{0.099}\\ \textcolor{codepurple}{-}\end{tabular} & \begin{tabular}[c]{@{}c@{}}\textcolor{blue}{0.233}\\ \textcolor{codegreen}{0.380}\\ \textcolor{codepurple}{-}\end{tabular} & \begin{tabular}[c]{@{}c@{}}\textcolor{blue}{-}\\ \textcolor{codegreen}{0.374}\\ \textcolor{codepurple}{-}\end{tabular}   & \begin{tabular}[c]{@{}c@{}}\textcolor{blue}{-}\\ \textcolor{codegreen}{0.371}\\ \textcolor{codepurple}{-}\end{tabular}   \\[22pt]
PSCC-Net                                                            & \begin{tabular}[c]{@{}c@{}}\textcolor{blue}{0.289}\\ \textcolor{codegreen}{0.580}\\ \textcolor{codepurple}{0.681}\end{tabular} & \begin{tabular}[c]{@{}c@{}}\textcolor{blue}{0.292}\\ \textcolor{codegreen}{0.589}\\ \textcolor{codepurple}{-}\end{tabular} & \begin{tabular}[c]{@{}c@{}}\textcolor{blue}{0.176}\\ \textcolor{codegreen}{0.478}\\ \textcolor{codepurple}{0.359}\end{tabular} & \begin{tabular}[c]{@{}c@{}}\textcolor{blue}{0.157}\\ \textcolor{codegreen}{0.428}\\ \textcolor{codepurple}{0.422}\end{tabular} & \begin{tabular}[c]{@{}c@{}}\textcolor{blue}{0.170}\\ \textcolor{codegreen}{0.358}\\ \textcolor{codepurple}{-}\end{tabular} & \begin{tabular}[c]{@{}c@{}}\textcolor{blue}{0.172}\\ \textcolor{codegreen}{0.353}\\ \textcolor{codepurple}{0.027}\end{tabular} & \begin{tabular}[c]{@{}c@{}}\textcolor{blue}{0.051}\\ \textcolor{codegreen}{0.101}\\ \textcolor{codepurple}{-}\end{tabular} & \begin{tabular}[c]{@{}c@{}}\textcolor{blue}{0.034}\\ \textcolor{codegreen}{0.068}\\ \textcolor{codepurple}{-}\end{tabular} & \begin{tabular}[c]{@{}c@{}}\textcolor{blue}{0.354}\\ \textcolor{codegreen}{0.577}\\ \textcolor{codepurple}{-}\end{tabular} & \begin{tabular}[c]{@{}c@{}}\textcolor{blue}{-}\\ \textcolor{codegreen}{0.085}\\ \textcolor{codepurple}{-}\end{tabular}   & \begin{tabular}[c]{@{}c@{}}\textcolor{blue}{-}\\ \textcolor{codegreen}{0.040}\\ \textcolor{codepurple}{-}\end{tabular}   \\[22pt]
TruFor                                                             &
\begin{tabular}[c]{@{}c@{}}\textcolor{blue}{0.381}\\ \textcolor{codegreen}{0.769}\\ \textcolor{codepurple}{0.740}\end{tabular} &
\begin{tabular}[c]{@{}c@{}}\textcolor{blue}{\underline{0.361}}\\ \textcolor{codegreen}{\underline{0.728}}\\ \textcolor{codepurple}{-}\end{tabular} &
\begin{tabular}[c]{@{}c@{}}\textcolor{blue}{\underline{0.271}}\\ \textcolor{codegreen}{0.734}\\ \textcolor{codepurple}{\underline{0.685}}\end{tabular} &
\begin{tabular}[c]{@{}c@{}}\textcolor{blue}{0.191}\\ \textcolor{codegreen}{0.520}\\ \textcolor{codepurple}{0.516}\end{tabular} &
\begin{tabular}[c]{@{}c@{}}\textcolor{blue}{\underline{0.245}}\\ \textcolor{codegreen}{\underline{0.461}}\\ \textcolor{codepurple}{-}\end{tabular} &
\begin{tabular}[c]{@{}c@{}}\textcolor{blue}{\textbf{0.408}}\\ \textcolor{codegreen}{\textbf{0.838}}\\ \textcolor{codepurple}{\underline{0.574}}\end{tabular} &
\begin{tabular}[c]{@{}c@{}}\textcolor{blue}{\underline{0.173}}\\ \textcolor{codegreen}{\underline{0.279}}\\ \textcolor{codepurple}{-}\end{tabular} &
\begin{tabular}[c]{@{}c@{}}\textcolor{blue}{\underline{0.080}}\\ \textcolor{codegreen}{\underline{0.159}}\\ \textcolor{codepurple}{-}\end{tabular} &
\begin{tabular}[c]{@{}c@{}}\textcolor{blue}{0.396}\\ \textcolor{codegreen}{0.646}\\ \textcolor{codepurple}{-}\end{tabular} &
\begin{tabular}[c]{@{}c@{}}\textcolor{blue}{-}\\ \textcolor{codegreen}{0.515}\\ \textcolor{codepurple}{-}\end{tabular}   &
\begin{tabular}[c]{@{}c@{}}\textcolor{blue}{-}\\ \textcolor{codegreen}{0.349}\\ \textcolor{codepurple}{-}\end{tabular}   \\[22pt]
FOCAL                                                              &
\begin{tabular}[c]{@{}c@{}}\textcolor{blue}{\textbf{0.459}}\\ \textcolor{codegreen}{\textbf{0.927}}\\ \textcolor{codepurple}{\textbf{0.950}}\end{tabular} &
\begin{tabular}[c]{@{}c@{}}\textcolor{blue}{\textbf{0.448}}\\ \textcolor{codegreen}{\textbf{0.904}}\\ \textcolor{codepurple}{-}\end{tabular} &
\begin{tabular}[c]{@{}c@{}}\textcolor{blue}{\textbf{0.289}}\\ \textcolor{codegreen}{\textbf{0.784}}\\ \textcolor{codepurple}{\textbf{0.739}}\end{tabular} &
\begin{tabular}[c]{@{}c@{}}\textcolor{blue}{\textbf{0.219}}\\ \textcolor{codegreen}{\textbf{0.596}}\\ \textcolor{codepurple}{\textbf{0.589}}\end{tabular} &
\begin{tabular}[c]{@{}c@{}}\textcolor{blue}{\textbf{0.298}}\\ \textcolor{codegreen}{\textbf{0.626}}\\ \textcolor{codepurple}{-}\end{tabular} &
\begin{tabular}[c]{@{}c@{}}\textcolor{blue}{\underline{0.281}}\\ \textcolor{codegreen}{\underline{0.576}}\\ \textcolor{codepurple}{\textbf{0.581}}\end{tabular} &
\begin{tabular}[c]{@{}c@{}}\textcolor{blue}{\textbf{0.177}}\\ \textcolor{codegreen}{\textbf{0.355}}\\ \textcolor{codepurple}{-}\end{tabular} &
\begin{tabular}[c]{@{}c@{}}\textcolor{blue}{\textbf{0.155}}\\ \textcolor{codegreen}{\textbf{0.311}}\\ \textcolor{codepurple}{-}\end{tabular} &
\begin{tabular}[c]{@{}c@{}}\textcolor{blue}{\underline{0.428}}\\ \textcolor{codegreen}{\underline{0.696}}\\ \textcolor{codepurple}{-}\end{tabular} &
\begin{tabular}[c]{@{}c@{}}\textcolor{blue}{-}\\ \textcolor{codegreen}{\underline{0.585}}\\ \textcolor{codepurple}{-}\end{tabular}   &
\begin{tabular}[c]{@{}c@{}}\textcolor{blue}{-}\\ \textcolor{codegreen}{\underline{0.429}}\\ \textcolor{codepurple}{-}\end{tabular}   \\ 
\bottomrule
\end{tabular}%
 }
\caption{Localization performance in terms of the mean weighted F1 score (F1$^{v_1}_w$) in popular datasets, for \textcolor{blue}{original dataset with tampered and pristine images}, \textcolor{codegreen}{original dataset with only tampered images} and \textcolor{codepurple}{only tampered images through \emph{Facebook}}. In \textbf{bold}, the highest score in each dataset, and \underline{underlined}, the second highest one.}
\label{tab:localization_f1_popular_v1}
\end{table}

\begin{table*}[]
\centering
\resizebox{\textwidth}{!}{%
\begin{tabular}{lccccccccc}
\toprule
Method           & Columbia & \begin{tabular}[c]{@{}c@{}}Columbia\\ WebP\end{tabular} & \begin{tabular}[c]{@{}c@{}}Casia 1.0\\ SP\end{tabular} & \begin{tabular}[c]{@{}c@{}}Casia 1.0\\ CM\end{tabular}  & Coverage & DSO-1 & Korus & \begin{tabular}[c]{@{}c@{}}Korus\\ WebP\end{tabular} & \begin{tabular}[c]{@{}c@{}}AutoSplice\\ 100\end{tabular} \\ 
\midrule
& & & & & & & & & \\[-8pt]
Noisesniffer     & 0.395 & 0.168   & 0.376 & 0.418           & 0.413    & 0.628 & \textbf{0.677} & 0.629     & 0.406         \\[10pt]
Zero             & 0.000   & 0.135 & 0.122 & 0.004            & 0.022    & 0.288 & 0.018 & \underline{0.836}     & 0.098         \\[10pt]
EXIF & 0.663   & 0.663 & \underline{0.539} & 0.537          & \textbf{0.645}    & \underline{0.655} & \underline{0.667} & \textbf{0.667}     & 0.761         \\[10pt]
PSCC-Net          & \underline{0.672}  & \underline{0.706}   & 0.516 &\underline{0.551}            & 0.292    & 0.199 & 0.579 & 0.306     & \underline{0.839}         \\[10pt]
TruFor           & \textbf{0.905}   & \textbf{0.849}  & \textbf{0.761} & \textbf{0.649}            & \underline{0.580}    & \textbf{0.857} & 0.488 & 0.424     & \textbf{0.865}         \\ 
\midrule
\end{tabular}%
}
\caption{Detection performance in terms of the dataset-level weighted F1 score (F1$^{v_2}_w$). The values correspond to the evaluation on tampered and untampered images of each dataset. In \textbf{bold}, highest value in each dataset, and \underline{underlined} second highest value in each dataset.}
\label{tab:detection_popular_F1}
\end{table*}

\section{Conclusions}
\label{sec:conclusions}
We presented \textit{PhotoHolmes}, a novel \textit{Python} library for forgery detection in digital images. The proposed library comprises a wide array of methods, datasets, and metrics that, when combined through the \textit{Benchmark} module, allows the user to do comparative studies of state-of-the-art methods. The library also provides a CLI that allows the user to run the included methods through suspicious images and save the outputs to localize the areas that are possibly forged. Given the design principles used, we aim to keep extending \textit{PhotoHolmes} by adding new methods, datasets, and metrics by building a community that contributes to it.

\bmhead{Acknowledgements}


The experiments presented in this paper were carried out using ClusterUY (site: https://cluster.uy)


\section*{Declarations}


\begin{itemize}
\item Funding: This research received no specific grant from any funding agency in the public, commercial, or not-for-profit sectors.
\item Competing interests: The authors have no conflicts of interest to declare.
\item Code, data and materials availability: Everything is publicly available in the library's GitHub.
\item Authors' contribution: J.O, R.P, J.S, J.U implemented the computational framework of the library. All of the authors contributed to the analysis of the results and writing of the manuscript.
\end{itemize}







\newpage
\begin{appendices}



\section{Datasets (Sec. \ref{subsec:datasets})}
\label{app:datasets}
In this section, we provide a brief description of the datasets included in the first release of \textit{PhotoHolmes}. 

\paragraph{Columbia.}
This dataset contains spliced images, which are not realistic at all and could be easily detected by semantic evaluation. This means that just by looking at the image and considering the context, a person can identify the suspicious area. One could argue that detecting forgeries of this type does not add value to a method, as they can be easily identified by the human eye. However, the importance of this dataset lies not only in its popularity but also in the fact that it has its version through different social networks~\cite{Wu2022}. With the correct metrics, it allows for the quantification of how well or poorly a method can generalize \textit{in the wild} forgeries, especially in the context of the different processing an image undergoes when uploaded to any social network. 

\paragraph{Casia 1.0.}
This dataset contains both splicing and copy move forgeries which are not so easy to identify to the naked eye and are JPEG compressed. It also has its version through different social networks~\cite{Wu2022} which allows the same analysis as Columbia on top of being spliced and copy move forgeries JPEG compressed.

\paragraph{Coverage.}
It is the most popular dataset for evaluating copy-move forgeries. The images in this dataset are uncompressed, and the pristine images consistently feature a repetition of a certain object. For the forged images, one of these objects is cut and pasted elsewhere, with the pasted object sometimes easily located and other times not. This dataset helps determine whether a method merely searches for similar parts within the image to detect a copy-move forgery or if it looks for inconsistencies in traces, such as the demosaicing grid.

\paragraph{DSO-1.}
DSO-1 is a dataset that contains spliced images in which the subject used for the splicing are humans. At first glance, the splices are hard to catch, however most of the times, doing a semantic evaluation regarding the light shows which subject is spliced. This dataset is of PNG images and it has its version through different social networks~\cite{Wu2022}.

\paragraph{Korus.}
The Korus dataset is also named realistic tampering. As the title suggests, this dataset contains forgeries that are almost impossible to detect through semantic evaluation. It has uncompressed images containing splicing copy move and object removal.

\paragraph{AutoSplice.}
This novel dataset is unique as it incorporates generative inpainting. Jia et al.~\cite{jia2023autosplice} introduce the utilization of \textit{DALL-E2} to generate forged images guided by a text prompt. These images are JPEG compressed, and the dataset includes variations with three JPEG quality factors: 100, 90, and 75. This diversity facilitates the quantification of how well methods can handle varying degrees of JPEG compression.

\paragraph{Trace.}
In Trace, the forged and pristine regions differ only in the traces left behind by the imaging pipeline. The concept involves selecting a raw image and processing it using two distinct imaging pipelines. The results are then merged, forming a single image with two areas, each corresponding to one of the two pipelines. The merging of these images is accomplished using a mask. An important consideration regarding Table~\ref{tab:localization_F1} is the fact that it was done in a subsampled version of the Trace dataset that will be made available with the publication of \textit{PhotoHolmes}.

\section{Methods (Sec. \ref{subsec:methods})} 
\label{app:methods}
In this section, we provide a brief description of the forensic methods implemented in the first release of \textit{PhotoHolmes}. 

\paragraph{Adaptive CFA Net.} This research paper presents an innovative approach to automatically detect suspicious regions in potentially forged images. The method uses a Convolutional Neural Network (CNN) to identify inconsistencies in image mosaics, specifically targeting the artifacts left by demosaicing algorithms. Unlike many blind detection neural networks, this approach does not require labeled training data and can adapt to new, unseen data quickly. The authors provide two sets of weights, first a pretrained version trained in a dataset that contained different types of demosaicing and then, a jpeg version that was trained with the same database as the other case but after it was compressed with a quality factor of 95. 

\paragraph{Noisesniffer.} The method exploits the consequences that the acquisition pipeline have on the noise model of a digital image. It estimates an stochastic model for said noise and detects noise anomalies using an a-contrario approach evaluating the number of false alarms (NFA). In order to get the suspected region of the forgery the authors of Noisesniffer use a region growing algorithm that detects the anomalous region according to the evaluation of the NFA. 

\paragraph{Zero.} The method detects JPEG compression as well as its grid origin. This method can be applied globally to identify a JPEG compression, and also locally to identify image forgeries when misaligned or missing JPEG grids are found. This allows image forensics to be applied, by identifying anomalies in the grid encountered locally with respect to the main grid detected.

\paragraph{DQ.} This method focuses on JPEG images and detects tampered regions by examining the double quantization effect hidden among the discrete cosine transform (DCT) coefficients.

\paragraph{CAT-Net.} CAT-Net is an end-to-end fully convolutional neural network designed to detect compression artifacts in images. CAT-Net combines both RGB and DCT streams, allowing it to simultaneously learn forensic features related to compression artifacts in these domains. Each stream considers multiple resolutions to deal with the various shapes and sizes of the spliced objects.

\paragraph{Splicebuster.} Splicebuster is a method based on finding anomalies in an image's residual, obtaining a set of features from the co-ocurrence matrix of a local estimation of the residual, which is assumed homogenous in the absence of forgery, and contain anomalies in the presence of such. This last assumption is explained through the fact that the image residual (from which the method's pipeline begins) is a reasonable estimation of the image's noise, which should have the same model in images of the same camera and different in other cases.

\paragraph{EXIF as Language.} EXIF as Language is a deep learning method based on a learned embedding between image patches and EXIF metadata. It is a joint embedding that makes a correspondence between image patch and EXIF metadata. The proposed method, converts the EXIF metadata to text and then it process it with a transformer and the patch is processed with a Patch encoder based on the ResNet architecture. The joint embedding is trained with contrastive learning. The authors provide two types of localization results, first a heatmap using the mean shift algorithm and then a mask using normalized cuts as clustering method. 


\paragraph{PSCC-Net.} PSCC-Net is an end-to-end fully convolutional neural network. It consists of a neural network that using a coarse to fine approach returns a mask locating forgeries in the input image. The method also returns an answer to the detection problem by returning a label that indicates whether the image was manipulated or not.

\paragraph{TruFor.} The paper presents a novel approach to detect and localize image forgeries. The method extracts both high-level and low-level features through a transformer-based architecture that combines the RGB image and a learned noise-sensitive fingerprint. The forgeries are detected as deviations from the expected regular pattern that characterizes a pristine image. On top of a pixel-level localization map and a whole-image integrity score, the method outputs a reliability map that highlights areas where the localization predictions may be error-prone, reducing false-alarms.

\paragraph{FOCAL.} FOCAL is based on a paradigm of contrastive learning and unsupervised clustering for the image forgery detection. It utilizes pixel-level contrastive learning to supervise the high-level forensic feature extraction in an image-by-image manner and employs an on-the-fly unsupervised clustering algorithm (instead of a trained one) to cluster the learned features into forged/pristine categories, further suppressing the cross-image influence from training data.

\section{Metrics (Sec.~\ref{subsec:metrics})}
\label{app:metrics}

\paragraph{True Positive Rate (TPR) or Recall.} It is the ratio between the True Positives and all the Positive (in ground truth) labelled data. Following a probabilistic interpretation, this would be the probability that a Positive labelled data be predicted as such.

\begin{equation*}
    TPR = \frac{TP}{P} = \frac{TP}{TP + FN}
\end{equation*}

\paragraph{False Positive Rate (FPR).} It is the ratio between the False Positives and all the Negative (in ground truth) labelled data.

\begin{equation*}
    FPR = \frac{FP}{N} = \frac{FP}{TN + FP}
\end{equation*}

\paragraph{Precision.} It is the ratio between the True Positives and all the predicted Positive.

\begin{equation*}
    \text{Precision} = \frac{TP}{TP + FP}
\end{equation*}

\paragraph{F1.} The F1 score is the harmonic mean of the precision and recall.

\begin{equation*}
    \text{F1} = \frac{2 TP}{2 TP + FN + FP}
\end{equation*}

\paragraph{Matthews correlation coefficient (MCC).} The MCC metrics that measures the quality of a binary prediction that can be used both for the classification task as for the localization task~\cite{MATTHEWS1975442}. The MCC is a value that goes from -1 (worst) to 1 (best).
\begin{equation*}
\begin{split}
    &MCC = \\
    &\frac{TP \times TN - FP \times  FN}{\sqrt{(TP + FP)(TP+FN)(TN+FP)(TN+FN)}}
    \label{C3:eq:MCC}
\end{split}
\end{equation*}

The MCC is often regarded as a measure of the quality of a confusion matrix. The difference between
this metric and precision / recall is that the MCC takes into account both true and false, positive 
and negative rates, with some authors \cite{Chicco2020} crowning it as the best binary classification
metric.

\paragraph{Intersection over Union (IoU).} Intersection over Union, also known as Jaccard Index, is a metric that measures the quality of a spatial prediction. In our case, the
prediction is the 2D forgery mask, but it is often used for bounding boxes in image segmentation tasks. 
This metric will allow us to measure the localization quality of a method.

\begin{equation*}
    IoU = \frac{TP}{TP + FP + FN}
    \label{C3:eq:IoU}
\end{equation*}

\paragraph{ROC.} The receiver operating characteristic (ROC) curve is a common metric used in binary classification. It is very similar to the precision/recall curve, but instead of plotting precision versus recall, the ROC curve plots the true positive rate (TPR) against the false positive rate (FPR) 

\paragraph{AUROC and mAUROC.} One way to compare classifiers using the ROC curve is to measure the area under the curve, this measurement is called AUROC. A perfect classifier will have a AUROC equal to 1, whereas a purely random classifier will have a AUROC equal to 0.5. 

The AUROC implemented by \textit{Torchmetrics} is done over the full dataset. However, the mean AUROC (mAUROC) is also implemented. The mean is done by varying the threshold in a single image, getting that ROC and then the AUROC from that, the aggregation is done by doing the mean of every AUROC.

\paragraph{Thresholding heatmaps.} Many of the aforementioned metrics assume that the prediction is a binary result; however, some methods generate a heatmap instead of a binary output. For instance, metrics like F1 score, MCC score, and IoU are designed for binary outputs. One approach to handle heatmaps is to binarize them using a threshold, often set to $0.5$ by default. Alternatively, an optimal threshold can be determined to maximize a specific metric on a chosen dataset.

\paragraph{Weighted metrics.} The authors of~\cite{Noisesniffer} and~\cite{bammey2021nonsemantic} present another solution by proposing the definition of weighted metrics. This allows for a comparison between methods that provide heatmaps and those that yield binary outputs. Gardella et al.~\cite{Noisesniffer} and Bammey et al.~\cite{bammey2021nonsemantic} suggest interpreting the heatmap at each pixel as the probability of the pixel being forged. With this perspective, they define weighted true positives, weighted false positives, weighted true negatives, and weighted false negatives as follows:

\begin{equation*}
    TP_w = \sum_xH(x)M(x)
    \label{C3:eq:weightedTP}
\end{equation*}

\begin{equation*}
    FP_w = \sum_x(1-M(x))H(x)
    \label{C3:eq:weightedFP}
\end{equation*}

\begin{equation*}
    TN_w = \sum_x(1-H(x))(1-M(x))
    \label{C3:eq:weightedTN}
\end{equation*}

\begin{equation*}
    FN_w = \sum_xM(x)(1-H(x))
    \label{C3:eq:weightedFN}
\end{equation*}

With the previous definitions, the F1 score, the MCC score and IoU defined in the previous subsections have their weighted version.

\end{appendices}


\bibliography{sn-bibliography, main, tools}

\end{document}